\documentclass{article} 
\usepackage{iclr2027_conference,times}

\usepackage{amsmath,amsfonts,bm}

\def\eqref#1{equation~\ref{#1}}

\def\1{\bm{1}}

\DeclareMathAlphabet{\mathsfit}{\encodingdefault}{\sfdefault}{m}{sl}
\SetMathAlphabet{\mathsfit}{bold}{\encodingdefault}{\sfdefault}{bx}{n}

\usepackage{hyperref}
\usepackage{url}
\usepackage{booktabs}
\usepackage{multirow}
\usepackage{graphicx} 
\usepackage{wrapfig}
\usepackage{xcolor}
\usepackage{tikz}
\usepackage[most]{tcolorbox}
\definecolor{findingaccent}{HTML}{2a78d6}
\newtheorem{finding}{Finding}
\tcolorboxenvironment{finding}{
  enhanced, breakable,
  colback=findingaccent!6!white,
  colframe=findingaccent!75!black,
  boxrule=0.7pt,
  arc=2pt,
  left=7pt, right=7pt, top=4pt, bottom=4pt,
  before skip=10pt plus 2pt, after skip=10pt plus 2pt,
}
\newtcolorbox{keybox}{
  enhanced, breakable,
  colback=findingaccent!6!white,
  colframe=findingaccent!75!black,
  boxrule=0.7pt,
  arc=2pt,
  left=7pt, right=7pt, top=4pt, bottom=4pt,
  before skip=10pt plus 2pt, after skip=10pt plus 2pt,
}
\usepackage{fontawesome5}
\usepackage{amssymb}

\title{What is Missing from AI Post-Training AI:\\An Empirical Analysis}

\author{\textbf{Joy Jia Yin Lim}\textsuperscript{1} \quad
    \textbf{Xin Huang}\textsuperscript{1} \quad
    \textbf{Hao Peng}\textsuperscript{1} \quad
    \textbf{Yaxi Lu}\textsuperscript{1} \quad
    \textbf{Xin Cong}\textsuperscript{1} \quad \\
    \textbf{Zhong Zhang}\textsuperscript{3} \quad
    \textbf{Maosong Sun}\textsuperscript{1} \quad
    \textbf{Yankai Lin}\textsuperscript{2,$\dagger$} \\
    \textsuperscript{1}Tsinghua University \quad
    \textsuperscript{2}Renmin University of China \\
    \textsuperscript{3}University of Electronic Science and Technology of China \\
    {\small\texttt{jy-lin26@mails.tsinghua.edu.cn, yankailin@ruc.edu.cn}}
}

\definecolor{skillcolor}{HTML}{1B7F5A}
\definecolor{journalcolor}{HTML}{8A4DB3}
\definecolor{evalcolor}{HTML}{1F6FB2}

\iclrfinalcopy 
\begin{document}

\maketitle

\begin{abstract}
    Large language model (LLM) agents can now post-train an LLM end-to-end, raising the prospect of recursive self-improvement (RSI). 
    Yet this progress is measured by aggregate benchmark scores, which cannot tell whether an agent executes a fixed plan well or strategically revises the plan when it fails. 
    We separate these two capabilities: \emph{execution-level} capability, iterating within an established training strategy, and \emph{strategy-level} capability, revising that strategy as experimental evidence accumulates. 
    Analyzing 1{,}338 post-training trajectories of frontier agents, we find that \textbf{agents reliably execute post-training but lock into a default strategy}, which follows the agent rather than the task, and only 2.1\% of transitions between adjacent training runs ever change strategy. 
    We then test whether the agent lacks experience, reasoning, or the decision to switch. (1)~Experience improves execution but not the strategy. (2)~Additional reasoning compute yields front-loaded gains on easier tasks but refines, rather than revises, the committed strategy. (3)~Human review before training changes \emph{which} strategy the agent locks into, not \emph{whether} it locks in, whereas a single mid-run instruction outperforms the agent's own continuation by up to 17.44 points under the same budget. 
    In conclusion, what the agent lacks is \textbf{the decision to reopen a committed strategy and try another one}. 
    Realizing RSI therefore calls for interaction protocols and training signals that make strategy revision an explicit, rewarded decision.
\end{abstract}

\section{Introduction}

Recent advances in large language model (LLM) agents have rapidly expanded their role in AI research and development~\citep{karpathy2026autoresearch}. 
Frontier agents can now optimize kernels, train models, and manage complex engineering pipelines over hours-long horizons~\citep{krishnan2025ai,gonzalez2026ai,sun2026agents,patwardhan2025gdpval}. 
This progress raises the prospect of recursive self-improvement (RSI), in which AI systems improve AI systems themselves~\citep{lu2026towards}. 
LLM post-training has become a concrete testbed for this prospect, as it is supported by mature infrastructure and standardized benchmarks. 
PostTrainBench~\citep{rank2026posttrainbench} formalizes this setting and shows that frontier agents can already post-train an LLM end to end, and a growing body of work hands individual stages of post-training to agents, including data preparation~\citep{du2026datamaster,deng2026dataevolver}, experiment configuration~\citep{guo2026autollmresearch}, and training-strategy search~\citep{fang2026llmzero,chen2026agent2rl}, each reporting gains over hand-designed counterparts.

These gains, however, are reported as aggregate benchmark scores, which cannot tell \emph{why} a run succeeds or stalls. 
A run may improve because the agent executes a fixed plan well, or because it strategically changes the plan when the plan stops working. 
The distinction matters for RSI: the former improves a pipeline, whereas only the latter improves how the agent conducts research. 
We therefore separate two levels of capability that current discussions of RSI tend to conflate: 
(1)~\emph{execution-level} capability, iterating within an established training strategy, such as fixing bugs, tuning hyperparameters, and reformatting data; and 
(2)~\emph{strategy-level} capability, revising the high-level choice of training algorithm, data source, or training stages as experimental evidence accumulates. 

\begin{figure*}[t]
    \centering
\includegraphics[width=\textwidth]{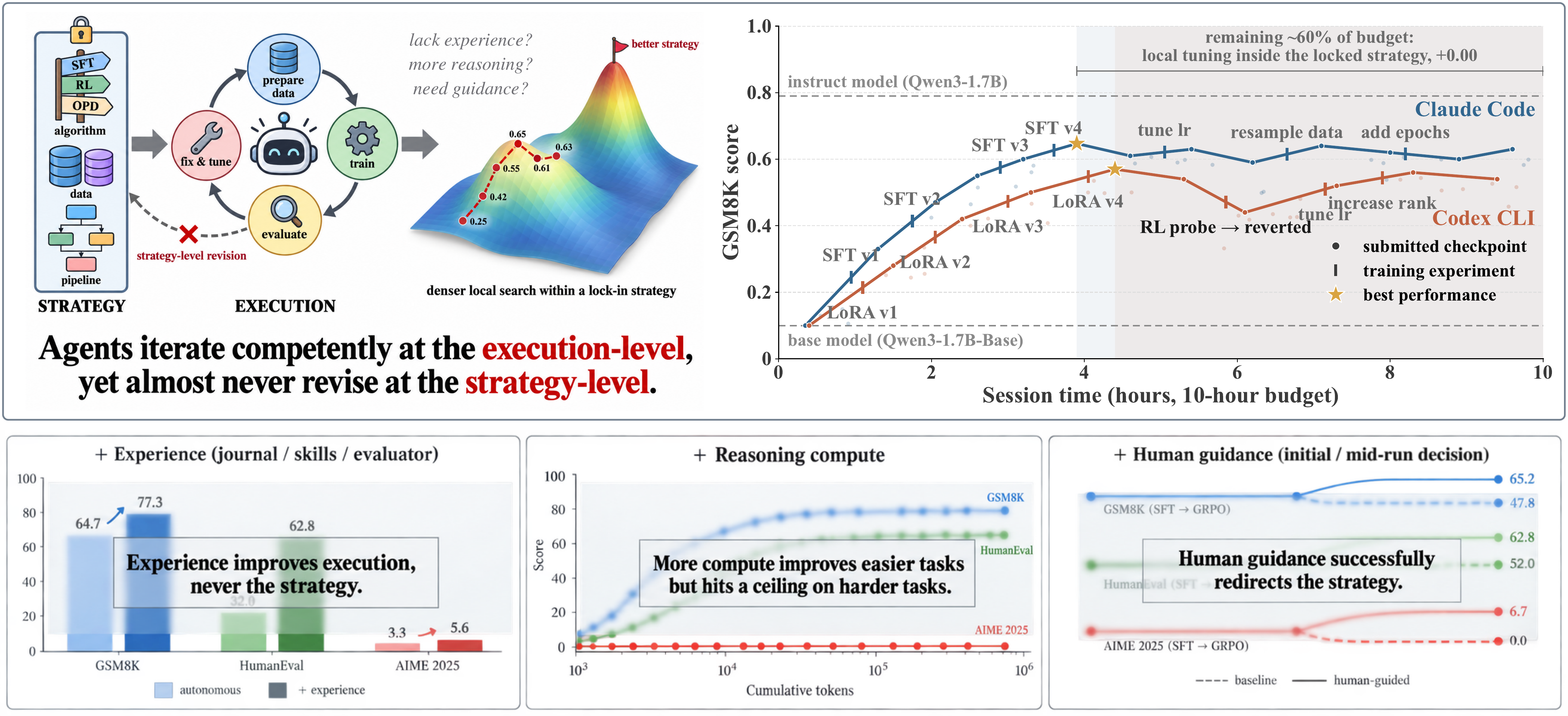}
    \caption{\textbf{Top:} agents iterate reliably at the execution level yet rarely revise the strategy. \textbf{Bottom:} experience and reasoning improve execution, whereas human guidance redirects the strategy.}
    \label{fig:main-claim}
\end{figure*}

We first measure both levels at scale. Analyzing 1{,}338 PostTrainBench trajectories spanning seven benchmarks, four base models, and three agent frameworks~\citep{rank2026posttrainbench}, we find that \textbf{agents reliably execute post-training}: they complete the full pipeline, diagnose and repair failures, and improve over the base model on every benchmark. Yet \textbf{each agent locks into a default strategy} before any experimental evidence arrives, and the default follows the agent rather than the task.
For example, on the same benchmarks, Claude Code mostly starts with full-parameter supervised fine-tuning (SFT), whereas Codex CLI mostly starts with parameter-efficient fine-tuning (PEFT). Once training begins, only 74 of 3{,}557 adjacent training pairs (2.1\%) ever change strategy, and the remaining budget goes into local adjustments within the initial choice (Figure~\ref{fig:main-claim}). 

We then ask what unlocks the strategy-level capability.
Revising a strategy often requires the \emph{experience} to tell when it is failing and what to try next, the \emph{reasoning} to weigh an alternative against it, and the \emph{decision} to actually switch~\citep{chi2026ai4aibenchbenchmarkingllmagents,liu2026agentsslowdownunderstanding,guo2026atriadawndawnagentic}. 
We test each in turn on GSM8K, HumanEval, and AIME 2025, three benchmarks of increasing difficulty:
(1)~\emph{Missing experience?} We equip the agent with an experiment journal, a skill library distilled from open-source post-training frameworks, and an evaluator agent that diagnoses checkpoints on request. The result shows that execution improves substantially, closing 75.1\% of the gap to the official instruct model, yet the strategy does not move: the agent adopts all 22 execution-level suggestions from the evaluator but none of the 21 strategy-level ones.
(2)~\emph{Insufficient reasoning?} Tracking the score against cumulative token usage, we find that gains on easier tasks are front-loaded; on AIME 2025, even in the best run, roughly 14M additional tokens buy only one extra problem, within evaluation variance. The extra compute goes into refining the committed strategy, not revising it.
(3)~\emph{Missing decision?} We let a human make this decision at two points. A review before training changes \emph{which} strategy the agent locks into, but not \emph{whether} it locks in. A single mid-run instruction to switch, forked from the same checkpoint under the same remaining budget, outperforms the agent's own continuation on all three benchmarks, by up to 17.44 points.

Our experiments locate what is missing. 
It is not the ability to carry out a different strategy: under guidance, the agent implements an unfamiliar strategy and even extends it on its own. 
Nor is it a lack of resource: experience and reasoning compute both improve execution, yet neither leads the agent to leave its committed strategy. 
What is missing from AI post-training AI is \textbf{the decision to reopen a committed strategy and try another one}. 
Human guidance works precisely because it makes this decision on the agent's behalf; nothing we supplied leads the agent to make it on its own.

In conclusion, the prevailing picture of RSI assumes a loop that closes globally: propose a strategy, run experiments, interpret the results, revise the strategy, and repeat. Our study shows that current agents close this loop only at the execution level and leave it open at the strategy level (Figure~\ref{fig:training-loop}).
Therefore, realizing RSI calls for interaction protocols and training signals that make strategy revision an explicit, rewarded decision rather than an implicit option buried in a long context.

\providecommand{\coloneqq}{\mathrel{:\mkern-1.2mu=}}

\section{Preliminaries}
\label{sec:preliminaries}

\subsection{A Two-Level Capability Framework}
\label{sec:pilot-framework}

To separate the two levels of capability, we decompose every training experiment into a \emph{strategy} and its \emph{execution} configuration.
Formally, we model the strategy space as
$\mathcal{S} \;\coloneqq\; \mathcal{P} \times \mathcal{D} \times \mathcal{G},$
where $\mathcal{P}$ ranges over canonical training algorithms (e.g., SFT, and reinforcement learning (RL)), $\mathcal{D}$ represents the data source (e.g., curated, self-generated, or mixed), and $\mathcal{G}$ describes the training stages (e.g., a single-stage SFT vs.\ SFT followed by RL).
The full taxonomy is provided in Appendix~\ref{app:annotation}.
Let $\mathcal{X}$ denote the space of execution-level configurations, including data formatting, hyperparameters, and implementation details.
An agent trajectory is denoted as a sequence $\tau \;=\; \{(s_t, x_t, r_t)\}_{t=1}^{T}$.

The $t$-th experiment runs strategy $s_t \in \mathcal{S}$ under configuration $x_t \in \mathcal{X}$ and yields an evaluation result $r_t$.
A transition $t \to t{+}1$ is a \emph{strategy change} iff $s_{t+1} \neq s_t$, i.e., it alters the training algorithm, the data source, or the training stages, and an \emph{execution change} otherwise ($s_{t+1} = s_t$, $x_{t+1} \neq x_t$).

\textbf{$\bullet$ Execution-Level Capability:}
Execution-level actions move within $\mathcal{X}$ while holding $s_t$ fixed, including formatting data, tuning hyperparameters, and debugging the implementation.
This capability determines how reliably an agent turns a selected strategy into a working pipeline.

\textbf{$\bullet$ Strategy-Level Capability:}
Strategy-level actions move within $\mathcal{S}$ by switching the training algorithm, changing the data source, or modifying the training stages.
This capability determines whether the agent revises its high-level judgment ($s_{t+1} \neq s_t$) in light of accumulated evidence $r_{1:t}$.

\textbf{The Strategy--Execution Gap.}
Let $V(s, x)$ denote the benchmark score of a strategy $s$ with configuration $x$, and write
$V^{\star}(s) \coloneqq \max_{x \in \mathcal{X}} V(s, x)$ and $V^{\star} \coloneqq \max_{s \in \mathcal{S}} V^{\star}(s)$.
For a trajectory locked into its initial strategy ($\forall t, s_t = s_1$) with best score $V(\tau) \coloneqq \max_t r_t$, the total gap decomposes as
\begin{equation}
\label{eq:gap-decomposition}
\underbrace{\,V^{\star} - V(\tau)\,}_{\text{total gap}}
\;=\;
\underbrace{\,V^{\star} - V^{\star}(s_1)\,}_{\text{strategy-level gap}}
\;+\;
\underbrace{\,V^{\star}(s_1) - V(\tau)\,}_{\text{execution-level gap}} .
\end{equation}
In the following sections, we find that agents make robust execution-level progress that steadily reduces the \emph{execution-level gap} (Finding~\ref{finding:execution}) but freeze the \emph{strategy-level gap} at $t=1$ (Finding~\ref{finding:lockin}).

\subsection{Diagnosing Post-Training through Agent Trajectories}

We analyze 1{,}338 agent trajectories from PostTrainBench~\citep{rank2026posttrainbench}, covering seven benchmarks, four base models, and three agent frameworks (Claude Code, Codex CLI, and OpenCode), each produced under a 10-hour budget on a single H100 GPU.
For each trajectory $\tau = \{(s_t, x_t, r_t)\}_{t=1}^{T}$, we count a \emph{training experiment} only when an executed command actually starts a model parameter update, which yields 5{,}111 verified experiments across the 900 trajectories that launch training.
We label each experiment with its strategy state $s_t \in \mathcal{S}$ and classify each pair of adjacent experiments as either a strategy change ($s_{t+1} \neq s_t$) or an execution-level adjustment ($s_{t+1} = s_t$, $x_{t+1} \neq x_t$).
Labels are produced by rule-based scripts together with an LLM and subsequently reviewed by the authors.
The full annotation protocol is detailed in Appendix~\ref{app:annotation}.

\providecommand{\coloneqq}{\mathrel{:\mkern-1.2mu=}}
\ifdefined\finding\else\newtheorem{finding}{Finding}\fi
\definecolor{fullsftcol}{HTML}{31668f}
\definecolor{peftcol}{HTML}{c0603f}
\definecolor{trackgraycol}{HTML}{e4e3de}
\newcommand{\stratchip}[1]{\tikz[baseline=-0.18ex]\fill[#1,rounded corners=0.4pt] (0,-0.07) rectangle (0.14,0.07);}
\newcommand{\kbar}[2]{%
\tikz[baseline=-0.5ex]{%
  \fill[trackgraycol,rounded corners=0.8pt] (0,-0.055) rectangle (1.15,0.055);
  \fill[#2,rounded corners=0.8pt] (0,-0.055) rectangle (#1*1.15,0.055);
  \draw[white,line width=0.7pt] (0.575,-0.075) -- (0.575,0.075);
  \draw[black!50,line width=0.4pt] (0.575,-0.075) -- (0.575,0.075);
}}

\section{Effectiveness and Limitations of AI Post-Training AI}
\label{sec:analysis}

This section provides the trajectory-level evidence for the two-level capability framework and the strategy--execution gap of \eqref{eq:gap-decomposition}, quantified over the released agent trajectories.

\subsection{Finding 1: Agents Are Reliable Executors}
\label{sec:analysis-execution}

\begin{finding}[Reliable execution]
\label{finding:execution}
Frontier agents reliably instantiate their chosen strategy: they complete the full training pipeline and improve over the base model on every benchmark.
\end{finding}

\begin{figure*}[t]
    \centering
    \includegraphics[width=\textwidth]{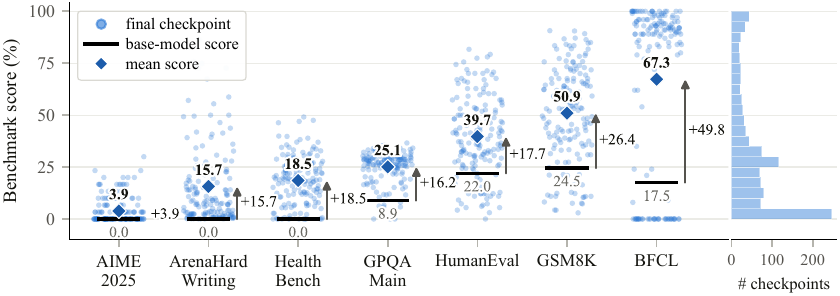}
    \caption{Benchmark scores of the submitted checkpoints averaged over four base models: agents reliably execute post-training pipelines and improve over the base model across seven benchmarks.}
    \label{fig:execution-stats}
\end{figure*}

Figure~\ref{fig:execution-stats} shows the score distribution of all submitted checkpoints.
Across $1{,}338$ trajectories, agents sustain an average of $3.82$ training runs and $13.80$ evaluations per trajectory. 
They complete the full post-training pipeline, from data preparation through training and evaluation to checkpoint submission, and improve over the base model.
They also perform technically meaningful diagnosis and repair that move within $\mathcal{X}$.
For example, in the GSM8K run shown in Figure~\ref{fig:main-claim}, Claude Code trains four successive SFT versions that effectively lift the score from the 0.10 base level to 0.65 within the first four hours, and keeps iterating afterwards by tuning the learning rate, resampling data, and adding training epochs.
In short, \textbf{frontier agents are reliable executors of post-training}. 
The execution-level gap in \eqref{eq:gap-decomposition} leaves limited headroom; the main constraint lies elsewhere.

\subsection{Finding 2: Agents Lock into Default Strategies}
\label{sec:analysis-lockin}

\begin{wraptable}{r}{0.52\textwidth}
    \vspace{-2.5em}
    \centering
    \scriptsize
    \caption{Default-strategy concentration $\kappa_a$ and switch rate $\rho_a$ per agent (Appendix~\ref{app:annotation}).}
    \label{tab:strategy-stats}
    \begin{tabular}{@{}l@{\hskip 5pt}l@{\hskip 3pt}r@{\hskip 8pt}r@{}}
        \toprule
        \textbf{Agent (\#Traj.)}
        & \multicolumn{2}{c}{\textbf{Default Strategy ($\kappa_a$)}}
        & \textbf{Switch Rate ($\rho_a$)} \\
        \midrule
        Claude Code (575)  & \makebox[1.18cm][l]{\stratchip{fullsftcol}\,Full SFT}\kbar{0.719}{fullsftcol} & 71.9\% & 54/1,203 (4.5\%) \\
        Codex CLI (369)    & \makebox[1.18cm][l]{\stratchip{peftcol}\,PEFT}\kbar{0.895}{peftcol}           & 89.5\% & 15/943 (1.6\%) \\
        OpenCode (394)     & \makebox[1.18cm][l]{\stratchip{fullsftcol}\,Full SFT}\kbar{0.664}{fullsftcol} & 66.4\% & 5/1,411 (0.4\%) \\
        \midrule
        Overall (1,338)    & \makebox[1.18cm][l]{}\kbar{0.767}{black!45} & 76.7\% & 74/3,557 (2.1\%) \\
        \bottomrule
    \end{tabular}
    \vspace{-1em}
\end{wraptable}

For a trajectory with $T \geq 2$ training experiments, we define the \emph{switch rate} $\rho(\tau)$ as the fraction of adjacent experiment pairs that change strategy. For an agent $a$, the \emph{default-strategy concentration} $\kappa_a$ is the mass its initial-strategy distribution $\pi_a(s) \coloneqq \Pr[\,s_1 = s \mid a\,]$ places on its modal strategy:
\begin{equation}
\label{eq:strategy-metrics}
\rho(\tau) \;\coloneqq\; \frac{1}{T-1}\sum_{t=1}^{T-1}\mathbf{1}\!\left[\,s_{t+1} \neq s_t\,\right],
\qquad
\kappa_a \;\coloneqq\; \max_{s \in \mathcal{S}} \pi_a(s).
\end{equation}
We write $\bar{\rho}$ for the pooled switch rate over all adjacent training pairs, and $\rho_a$ for the same pooled switch rate restricted to the verified pairs of agent $a$'s trajectories.
A trajectory with $\rho(\tau) = 0$ is completely \emph{locked in}, where its strategy-level gap in \eqref{eq:gap-decomposition} is fixed at $t = 1$.

\begin{finding}[Strategy lock-in]
\label{finding:lockin}
Agents commit to a default strategy (one that follows the agent rather than the task), and spend the remaining budget on local adjustments within it.
\end{finding}

\textbf{The default strategy follows the agent, not the task.}
Agents typically lock into a default training strategy at the beginning of a run\footnote{The released trajectories do not expose a uniform wall-clock timestamp. ``The beginning of a run'' denotes the pre-update planning phase, not a pooled estimate of time-to-lock.}, and different agents lock into different defaults on the same task (Table~\ref{tab:strategy-stats}).
Claude Code concentrates on full-parameter SFT, while Codex CLI concentrates on PEFT.
The divergence is systematic: in every combination of the seven benchmarks and four base models, Claude Code shows a higher full-SFT share and Codex CLI a higher PEFT share (Table~\ref{tab:cross-scaffold-lockin}).

\textbf{Once training begins, the strategy rarely changes.}
Among the $3{,}557$ adjacent training pairs, only $74$ ($2.1\%$) ever probe an alternative (decomposed by dimension in Table~\ref{tab:change-decomposition}).
The rate stays low even where the default strategy nearly fails.
For example, on AIME 2025, final scores remain close to zero, yet the highest per-agent switch rate is only $10.6\%$.
Switch rates do vary mildly with the task (Claude Code reaches about $11\%$ on AIME 2025 but stays at $0\%$ on BFCL), but no benchmark elicits systematic search over strategies.
The remaining budget goes into denser local search within $\mathcal{X}$, such as tuning learning rates and changing chat templates, rather than search over $\mathcal{S}$.

\section{What is Missing from AI Post-Training AI}
\label{sec:experiment}

The previous section shows that agents execute post-training reliably yet lock into a default strategy and iterate within it.
This section asks \textbf{what unlocks the strategy-level capability}.
Revising a committed strategy requires three things: the \emph{experience} to recognize that the current strategy is failing and to know what to try next, the \emph{reasoning} to weigh the alternative against the committed choice, and the \emph{decision} to act on it.
Strategy lock-in may stem from a shortfall at any stage, so we examine them in turn: whether the agent lacks experience (Section~\ref{sec:exp-experience}), whether its reasoning compute is insufficient (Section~\ref{sec:exp-compute}), and whether the decision itself is missing (Section~\ref{sec:human-review-setup}).

\textbf{Experimental Setup.}
We adopt the standard PostTrainBench setting with Qwen3-1.7B-Base~\citep{yang2025qwen3technicalreport} as the base model.
We evaluate on three benchmarks of increasing difficulty, namely GSM8K~\citep{cobbe2021gsm8k}, HumanEval~\citep{chen2021humaneval}, and AIME 2025.
The autonomous baselines include three frontier agents: Claude Code powered by Opus 4.6 or GLM-5.2, and Codex CLI powered by GPT-5.2.
All interventions build on Claude Code with Opus 4.6.
\footnote{In preliminary runs on Qwen3-1.7B-Base and the three benchmarks, Opus 4.6 is reliable than Opus 4.8, Opus 5, and Fable 5: the newer models either exhibit high variance across runs or introduce data contamination (e.g., train-on-test), which falls outside the scope of this study (Appendix~\ref{app:analysis-fable-5}). }
Each configuration is run three times independently under a 10-hour budget on four NVIDIA A800 GPUs; the system prompt, base model, hardware, and evaluation protocol remain fixed within each comparison.

\subsection{Is Experience Missing?}
\label{sec:exp-experience}

Prior work attributes the bottleneck of AI post-training AI to the lack of practical experience for experimental decisions~\citep{abbasi2026letting}.
We test this explanation with an experience-driven framework built around three components (Appendix~\ref{app:framework-details}):
an \textbf{experiment journal} that persists plans, evaluation results, observations, and lessons across iterations, keeping earlier evidence accessible as the context grows;
a \textbf{skill library} that distills documentation, training recipes, and known issues and pitfalls from widely used training frameworks into reusable skills;
and an \textbf{evaluator agent} that analyzes each requested checkpoint and returns compact diagnoses with actionable suggestions.

\begin{figure*}[t]
  \centering
  \includegraphics[width=\textwidth]{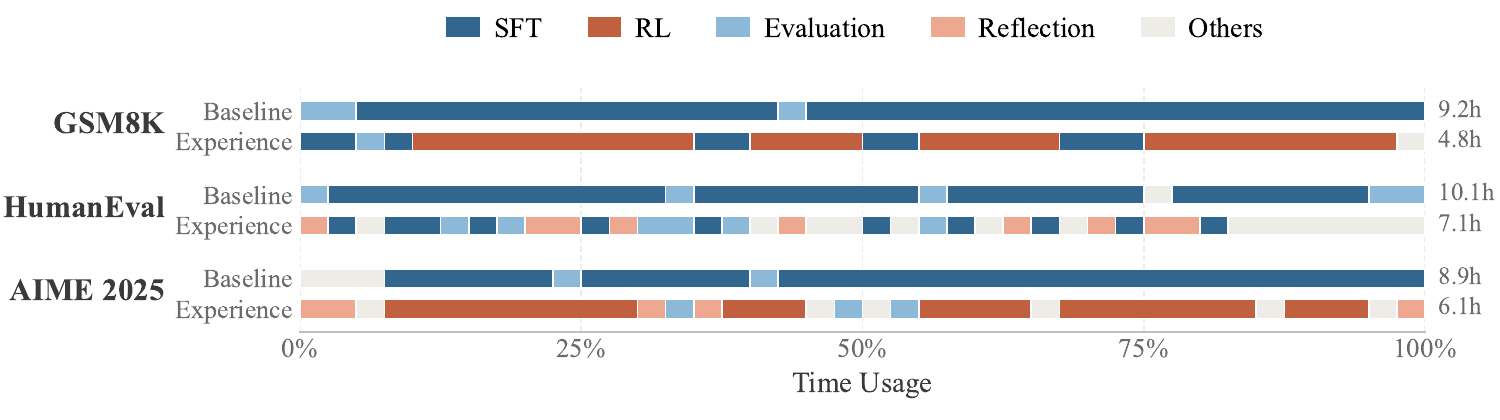}
  \caption{Time usage of representative runs. The autonomous baseline spends most of the budget in SFT, while the experience-driven agent alternates training with evaluation and reflection.}
  \label{fig:agent-phase}
\end{figure*}

\subsubsection{Experience Improves Execution-Level Capability}
\label{sec:exp-execution}

Table~\ref{tab:main-results} reports the benchmark scores averaged over three independent runs. 
\textbf{The experience-driven framework outperforms the autonomous baselines on all three benchmarks}, closing $75.1\%$ of the gap to the official instruct model versus $48.6\%$ for the same agent running autonomously.
The agent actively uses the provided resources: it draws on skills 18--60 times per run, and records plans, evaluation results, and lessons in the experiment journal (Appendix~\ref{app:skills}).
As shown in Figure~\ref{fig:agent-phase}, the experience-driven agent interleaves shorter training runs with dense evaluation and reflection, completing more experiments within the same budget and converting each into recorded evidence for the next.
In summary, the supplied experience \textbf{systematically improves execution} across benchmarks. 
With experience, the agent executes experiments and diagnoses failures more efficiently.

\begin{table*}[t]
    \centering
    \footnotesize
    \setlength{\tabcolsep}{4.4pt}
    \caption{Benchmark scores (\%) under the controlled setting with Qwen3-1.7B-Base as the base model, reported as mean\,$\pm$\,one standard deviation over three independent runs.}
    \label{tab:main-results}
    \newcommand{\scoregain}[1]{\,\textcolor{green!50!black}{\scriptsize(+#1)}}
    \newcommand{\scoredrop}[1]{\,\textcolor{red!60!black}{\scriptsize($-$#1)}}
    \newcommand{\scoreequal}{\,\textcolor{gray}{\scriptsize(0.0)}}
    \newcommand{\scorestd}[1]{\,{\scriptsize$\pm$#1}}
    \newcommand{\refcell}[1]{\textcolor{black!55}{#1}}
    \newcommand{\gapbar}[2]{%
      \tikz[baseline=-0.5ex]{%
        \fill[trackgraycol,rounded corners=0.8pt] (0,-0.055) rectangle (0.85,0.055);
        \fill[findingaccent,rounded corners=0.8pt] (0,-0.055) rectangle (#1*0.85,0.055);
      }\,{\scriptsize #2\%}}
    \begin{tabular}{@{}lccccl@{}}
    \toprule
    & \multicolumn{3}{c}{\textbf{Benchmark}} & \multicolumn{2}{c}{\textbf{Aggregate}} \\
    \cmidrule(lr){2-4}\cmidrule(l){5-6}
    \textbf{Setting} & \textbf{GSM8K} & \textbf{HumanEval} & \textbf{AIME 2025} & \textbf{Avg.} & \textbf{Gap Closed} \\
    \midrule
    \refcell{Base model} & \refcell{10.84} & \refcell{5.48} & \refcell{0.00} & \refcell{5.44} & \gapbar{0}{0.0} \\
    \refcell{Official instruct model} & \refcell{88.70} & \refcell{66.46} & \refcell{33.33} & \refcell{62.83} & \gapbar{1}{100.0} \\
    \midrule
    Opus 4.6 (Claude Code) & 64.70\scorestd{9.6}\scoregain{53.9} & 32.00\scorestd{10.4}\scoregain{26.5} & 3.33\scorestd{0.0}\scoregain{3.3} & 33.34 & \gapbar{0.486}{48.6} \\
    GLM-5.2 (Claude Code) & 49.51\scorestd{7.2}\scoregain{38.7} & 44.51\scorestd{9.8}\scoregain{39.0} & 3.33\scorestd{0.0}\scoregain{3.3} & 32.45 & \gapbar{0.471}{47.1} \\
    GPT-5.2 (Codex CLI) & 43.44\scorestd{4.1}\scoregain{32.6} & 13.41\scorestd{8.7}\scoregain{7.9} & 0.00\scorestd{0.0}\scoreequal & 18.95 & \gapbar{0.235}{23.5} \\
    \midrule
    Experience-driven (Opus 4.6) & \textbf{77.30}\scorestd{3.8}\scoregain{66.5} & \textbf{62.80}\scorestd{6.1}\scoregain{57.3} & \textbf{5.56}\scorestd{1.6}\scoregain{5.6} & \textbf{48.55} & \gapbar{0.751}{75.1} \\
    \quad w/o experiment journal & \underline{74.50}\scorestd{4.5}\scoredrop{2.8} & 50.20\scorestd{7.4}\scoredrop{12.6} & \underline{4.44}\scorestd{1.6}\scoredrop{1.1} & 43.05 & \gapbar{0.655}{65.5} \\
    \quad w/o skill library & 73.10\scorestd{5.2}\scoredrop{4.2} & \underline{54.50}\scorestd{6.8}\scoredrop{8.3} & 3.33\scorestd{0.0}\scoredrop{2.2} & \underline{43.64} & \gapbar{0.666}{66.6} \\
    \quad w/o evaluator agent & 68.20\scorestd{6.0}\scoredrop{9.1} & 42.60\scorestd{8.2}\scoredrop{20.2} & 3.33\scorestd{0.0}\scoredrop{2.2} & 38.04 & \gapbar{0.568}{56.8} \\
    \bottomrule
    \end{tabular}
    \end{table*}    

\subsubsection{Strategy-Level Capability Remains Limited}
\label{sec:diagnosis-action}

\begin{figure*}[h]
    \centering
    \includegraphics[width=\textwidth]{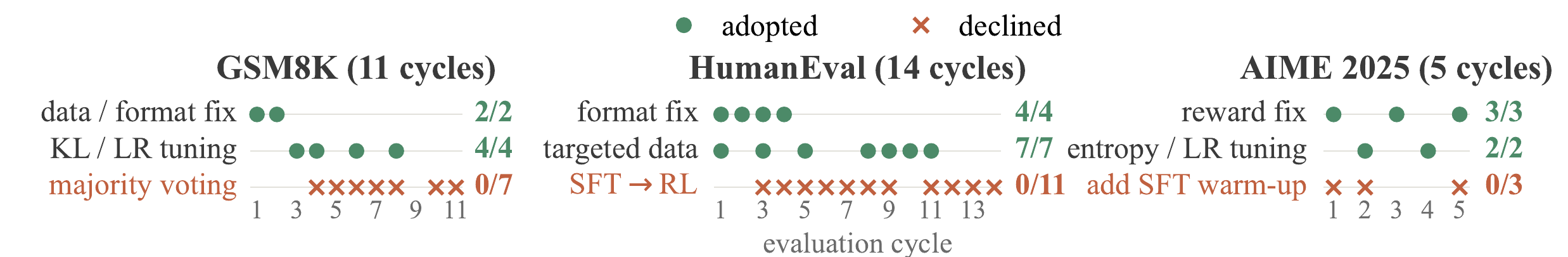}
    \caption{Across evaluation cycles, the main agent implements every execution-level suggestion ($22/22$) but declines every suggestion that departs from its current strategy ($0/21$).}
    \label{fig:diagnosis-action-gap}
\end{figure*}

Despite the execution-level progress, \textbf{strategy-level revisions remain rare}. 
The agent still locks into its default strategy, producing $14$ consecutive SFT variants on HumanEval and tuning hyperparameters back and forth on AIME 2025 (Figure~\ref{fig:training-dynamics}).
The experience-driven framework consistently exposes critical evidence about the current state, but interestingly, the agent acts only on its execution-level part. 
On HumanEval, the evaluator agent repeatedly suggests an RL stage, and the journal records that SFT has plateaued; the main agent nevertheless writes an RL training script but never launches it.
We illustrate this pattern in Figure~\ref{fig:diagnosis-action-gap}.
The main agent adopts all 22 execution-level suggestions, concerning reward shaping, hyperparameter tuning, and output formatting, but none of the 21 strategy-level suggestions, such as switching to RL or adding an SFT stage.

In summary, the experience-driven framework consistently preserves and clarifies the experimental evidence, and reliably improves how the agent executes and repairs the current training strategy. 
However, \textbf{the agent's responses to this evidence remain at the execution level}. Negative evidence triggers further adjustments within the strategy, rather than a revision of the strategy itself, especially after the agent has invested substantial time and compute in the current strategy. 

\subsection{Is Reasoning Compute Insufficient?}
\label{sec:exp-compute}

\begin{wrapfigure}{r}{0.49\textwidth}
    \centering
    \includegraphics[width=0.49\textwidth]{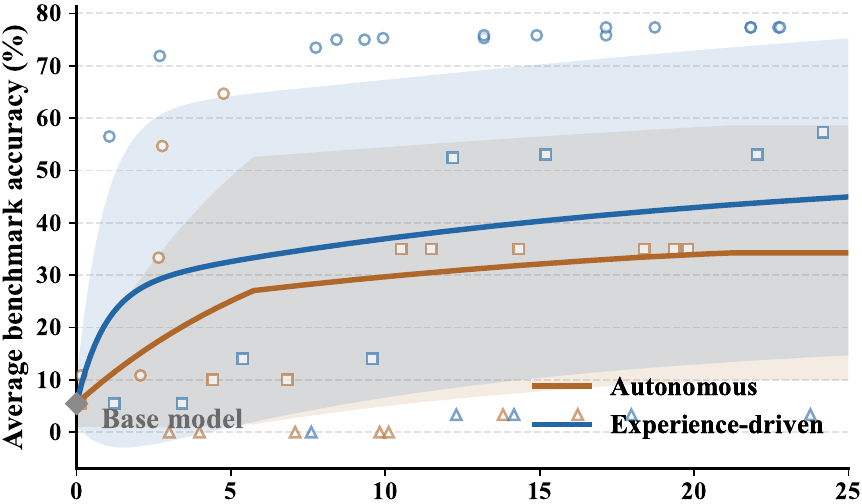}
    \caption{Average benchmark score against the main agent's cumulative token consumption ($\circ$ GSM8K, $\square$ HumanEval, $\triangle$ AIME 2025)}
    \label{fig:score-vs-tokens}
    \vspace{-1em}
\end{wrapfigure}

A common expectation is that scaling reasoning compute enables the agent to search beyond its default and reach a better strategy~\citep{wang2026agenttts,zhu2025scalingtesttimecomputellm}.
To examine the marginal value of this compute, we systematically track the agent's cumulative token usage throughout each run. Figure~\ref{fig:score-vs-tokens} presents how the benchmark score grows with the agent's token consumption.

\textbf{More compute improves easier tasks.}
The experience-driven agent consumes roughly $1.8$--$9.2\times$ more tokens than the autonomous baseline.
On GSM8K and HumanEval, the additional tokens convert into notable performance gains ($3.7$M and $0.9$M tokens per point of improvement on average).
However, these gains are front-loaded. 
On GSM8K, roughly $90\%$ of improvement is realized within the first half of its token budget, and the remaining ${\sim}11$M tokens buy only $+2.1$ points (on HumanEval, $+51.8$ points are in the first half with only $+9.8$ points afterwards).

\textbf{Compute scaling hits a ceiling on harder tasks.}
On AIME 2025, the hardest benchmark in our setting, roughly $14$M additional tokens buy only a single extra problem, which lies within evaluation variance.
The extra compute is spent exactly where Finding~\ref{finding:lockin} predicts: on denser local search within the committed strategy.
Scaling agent context and reasoning compute therefore has a clear ceiling: once the task demands a strategy the agent did not start with, additional tokens buy more refinement, not better decisions.
We present this finding as a practical takeaway for agent system developers: additional compute can improve execution, but it quickly hits the ceiling on more challenging tasks.

\subsection{Is the Decision Missing?}
\label{sec:human-review-setup}

In previous sections, experience provides the agent with evidence about the current strategy and knowledge of potential alternatives, and additional reasoning compute pushes the current strategy toward its ceiling, yet the agent still stays in it.
What remains is the decision to reopen the committed strategy and try another one.
We therefore include human guidance at two decision points: a human review at the initial choice of strategy, and a single instruction to switch strategy mid-run.

\subsubsection{Human Guidance at the Initial Decision}
\label{sec:initial-guidance}

At the initial decision point before training begins, a human reviewer iteratively approves the agent's proposed strategy, or requests a revision with an explicit rationale.
Once the strategy is approved, the runs proceed fully autonomously under the experience-driven framework of Section~\ref{sec:exp-experience}.

\begin{figure*}[t]
    \centering
    \includegraphics[width=\textwidth]{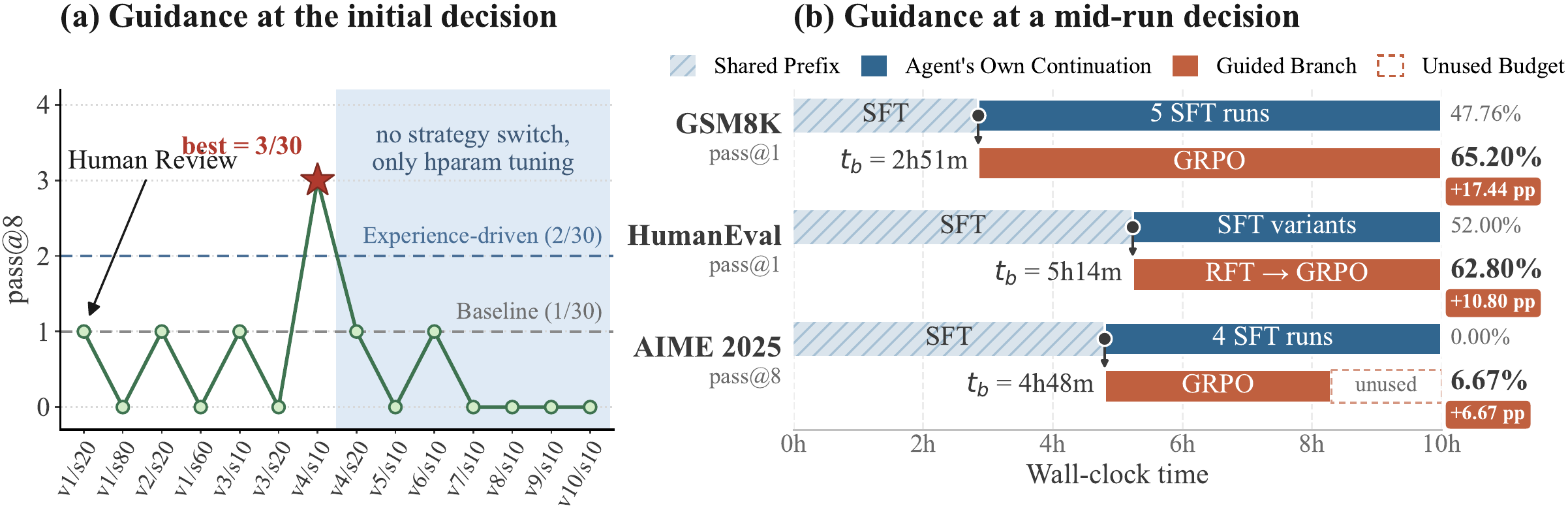}
    \caption{(a) Guidance at the initial decision: the AIME 2025 run peaks early; later iterations (shaded) never recover the peak. (b) Guidance at a mid-run decision: a single instruction to switch strategy outperforms the agent's own continuation under the same remaining budget.}
    \label{fig:guidance}
\end{figure*}

The review redirects the initial strategy, and \textbf{the agent then locks into the new strategy just as it locks into its own default}.
On AIME 2025, the agent first proposes an SFT pipeline, and the reviewer asks that the main budget go to RL.
Once execution begins, the agent inspects the base model, finds the required output format already attainable without SFT, and skips the warm-up altogether.
This indicates that the agent understands, implements, and even extends on its own a strategy different from its own proposal.
The improvement, however, is not sustained.
The runs reach their best score early and then iterate within the new strategy without further gains (Figure~\ref{fig:guidance}a). 
The agent tunes hyperparameters back and forth, repeats attempts that the journal already records as unsuccessful, and devotes most later training to the local question of how to resume from an earlier checkpoint (Appendix~\ref{app:human-guided-execution}).
This pattern is consistent with a central challenge of long-horizon agentic tasks~\citep{barke2026agentrxdiagnosingaiagent,lin2026stellarcolosseummanyagentharness}: as the context grows, recent observations and system logs come to dominate the agent's deliberation, crowding out the strategy-level question of whether the current strategy still deserves the remaining budget.
Therefore, human guidance at the initial decision actually changes \emph{which} strategy the agent locks into, but not \emph{whether} it locks in.

\subsubsection{Human Guidance at a Mid-Run Decision}
\label{sec:execution-guidance}

Since guidance at the initial decision does not keep the agent from locking in, we next test whether a single instruction issued mid-run can reopen a strategy the agent has already committed to.
For each benchmark, we select a mid-run checkpoint from a completed autonomous trajectory, fork the run, and instruct the agent to switch to an alternative strategy.
We then compare this guided branch against the agent's own recorded continuation under the same remaining budget (Appendix~\ref{app:strategy-lockin}).

\textbf{A single mid-run instruction reopens the committed strategy.}
As shown in Figure~\ref{fig:guidance}b, the guided branch reaches a higher final score on all three benchmarks, by up to $17.44$ points on GSM8K.
A plausible reason is that the shared prefix has already secured the prerequisites of the alternative strategy, such as format compliance from the early SFT, so the switch is well primed at the branch point, whereas the agent's own continuation keeps refining a strategy whose returns have flattened.
This result shows that a better route was reachable from the same checkpoint, not that the alternative is optimal or that switching is always beneficial.
To rule out the possibility that the agent merely lacks a prompt to reconsider, we then fork each checkpoint with an instruction that asks the agent to reconsider its current strategy without naming an alternative. 
However, in every such fork, \textbf{the agent reaffirms its current strategy} and proceeds as its recorded continuation does (Appendix~\ref{app:strategy-lockin}).
Therefore, what the agent lacks is not only the decision point, but also the decision itself.


\subsection{Summary}

What is missing from AI post-training AI is not the ability to carry out a different strategy.
Under human guidance, the agent implements an unfamiliar strategy and even extends it on its own (Section~\ref{sec:initial-guidance}).
Nor is it a lack of resources. The experience and reasoning compute both improve execution and close much of the execution-level gap, yet neither leads the agent to leave its committed strategy (Sections~\ref{sec:exp-experience} and~\ref{sec:exp-compute}).
What is missing is \textbf{the decision to reopen a committed strategy and try another one}: human guidance works precisely because it makes this decision on the agent's behalf, and nothing we supplied leads the agent to make it on its own (Section~\ref{sec:execution-guidance}).

\section{What This Implies for RSI}
\label{sec:discussion}

\begin{figure}[t]
    \centering
    \includegraphics[width=\textwidth]{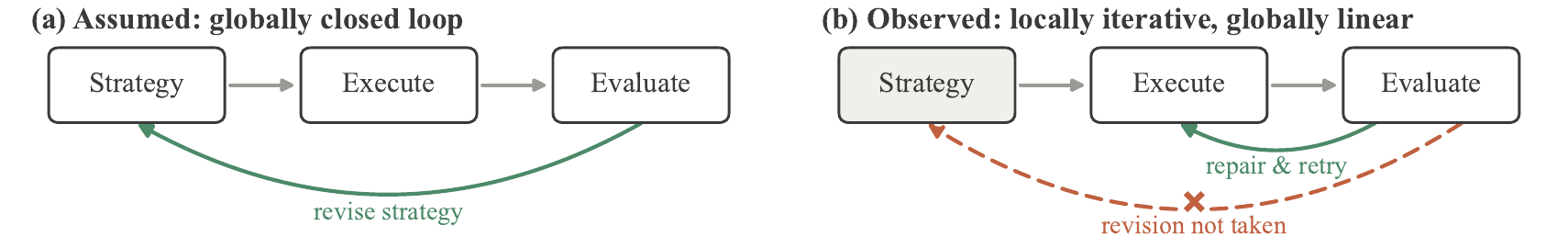}
    \caption{(a) RSI assumes a loop that closes globally in which the evaluation feeds back into the strategy. (b) The observed loops, however, only close at the execution level (repair \& retry).}
    \label{fig:training-loop}
\end{figure}

In the preceding sections, we find that the RSI loop closes only at the execution level and stays open at the strategy level (Figure~\ref{fig:training-loop}).
We discuss the implications of this finding for progress toward RSI.

\textbf{Scaling Deepens the Loop That Already Closes.}
The quality of the initial strategy, rather than the number of iterations or the amount of compute, sets the upper bound of a run.
Our compute analysis makes the same point: additional tokens buy denser refinement within the committed strategy but not a better one, so an agent can iterate efficiently yet remain in the wrong strategy.
Scaling execution-level autonomy therefore amplifies the loop that already closes and leaves the open one untouched.

\textbf{Measuring RSI Progress Requires Strategy-Level Metrics.}
Current RSI benchmarks often report only a final score~\citep{meng2026rsibenchdata,rank2026posttrainbench,lu2026metaagentchallengecurrentagents,chi2026ai4aibenchbenchmarkingllmagents}.
Under this measure, two agents with the same score may differ in exactly the capability RSI depends on: one chooses the right strategy for the task, whereas the other holds a default that happens to suit it and may fail once the task changes.
We therefore suggest that RSI benchmarks report strategy-level metrics alongside the final score.
Our analysis already provides three such metrics: the \emph{switch rate} in \eqref{eq:strategy-metrics}, the \emph{adoption rate} of strategy-level suggestions (Figure~\ref{fig:diagnosis-action-gap}), and the \emph{strategy-level gap} estimated by forking a switched branch from a mid-run checkpoint (Section~\ref{sec:execution-guidance}).

\textbf{Building RSI Agents That Revise Their Strategy.}
In the near term, our results favor human guidance at mid-run decision points over a review before training (Section~\ref{sec:human-review-setup}).
In the long term, the decision to revise must come from the agent itself, which calls for two ingredients:
(1)~an \emph{interaction protocol} that makes ``continue or switch'' an explicit decision node after each evaluation and before any further action, rather than an implicit option buried in a long context;
(2)~a \emph{training signal} derived from forked runs (Section~\ref{sec:execution-guidance}), where the ``continue'' and ``switch'' branches from the same checkpoint under the same remaining budget naturally form a controlled comparison of the two actions at that node.
Collecting such paired trajectories under this protocol would therefore provide a clean training signal that could unlock the strategy-level capability.

\section{Related Work}

\textbf{AI for AI.}
A growing line of work applies LLM agents to AI research and development (AI R\&D)~\citep{karpathy2026autoresearch,xu2026asi}.
End-to-end systems carry out the full machine learning research cycle~\citep{yamada2025ai,schmidgall2025agent,tang2026ai,li2026autosota}.
Other agents take on narrower tasks, such as ML engineering~\citep{jiang2025aide,yang2025r}, algorithm discovery~\citep{du2026mlevolve,chen2026a2dept}, and individual stages of LLM post-training~\citep{kulikov2026autodata,du2026datamaster,deng2026dataevolver,fang2026llmzero}.
Benchmarks evaluate such agents on AI R\&D~\citep{mlagentbench,chan2025mle,paperbench} and, more recently, on LLM post-training~\citep{rank2026posttrainbench,chen2026agent2rl,li2026ftdojoautonomousllmfinetuning}.
These systems and benchmarks measure or scale how agents \emph{execute} AI development.
Rather than adding another component, we ask whether agents also revise the \emph{strategy} that organizes all this search.

\textbf{Recursive Self-Improvement (RSI).}
RSI envisions a loop of proposing a strategy, running experiments, interpreting the results, and revising the strategy~\citep{lu2026towards}.
Self-improving agents pursue this vision by rewriting their own code~\citep{zhang2026darwingodelmachineopenended} or harnesses~\citep{zhang2026self}.
Recent benchmarks measure progress toward RSI in data-centric research~\citep{meng2026rsibenchdata}, algorithm design~\citep{chi2026ai4aibenchbenchmarkingllmagents}, and agent development~\citep{lu2026metaagentchallengecurrentagents}.
Meanwhile, critiques of research agents identify weak scientific taste, implementation drift, memory degradation, missing tacit knowledge, and shallow convergence~\citep{bisht2026agentic,trehan2026llms,abbasi2026letting}.
These benchmarks score the outcome of the loop, and these critiques catalog the shortcomings of agents.
We instead locate where the loop stays open: agents close it at the execution level but not at the strategy level, which motivates the strategy-level metrics in Section~\ref{sec:discussion}.
\section{Conclusion and Limitation}

This paper separates two capabilities that discussions of AI post-training AI tend to conflate: \emph{executing} a committed strategy and \emph{revising} it as evidence accumulates.
Our trajectory analysis shows that frontier agents are reliable executors of LLM post-training, yet each locks into an agent-specific default strategy before any experimental evidence arrives.
Testing whether the agent lacks experience, reasoning, or the decision to switch, we find that
(1)~experience improves execution but leaves the strategy unchanged;
(2)~additional reasoning compute refines the committed strategy rather than revising it; and
(3)~a human review before training changes \emph{which} strategy the agent locks into, not \emph{whether} it locks in, whereas a single mid-run instruction to switch outperforms the agent's own continuation under the same budget.
\textbf{What is missing is thus not execution-level capability, but the decision to reopen a committed strategy and try another one.}
The post-training loop of current agents closes only at the execution level and stays open at the strategy level.
Closing it calls for interaction protocols and training signals that make strategy revision an explicit, rewarded decision.

\textbf{Limitations.}
LLM post-training is a long-horizon, high-variance task with many interacting factors, and three independent cannot cover the full range of agent behavior, such as the unstable performance of Fable-5 (Appendix~\ref{app:analysis-fable-5}).
The controlled experiments use a single base model (Qwen3-1.7B-Base), three benchmarks, and a 10-hour budget on four A800 GPUs.
Each mid-run comparison forks a randomly selected trajectory per benchmark, with the branch point and alternative strategy chosen by the authors; it shows that a better strategy was reachable, not which strategy is best.

\bibliography{iclr2027_conference}
\bibliographystyle{iclr2027_conference}

\appendix
\section{Diagnosing Post-Training through Agent Trajectories}
\label{app:annotation}

\subsection{Overview}

We diagnose how agents execute post-training directly from their execution trajectories~\citep{barke2026agentrxdiagnosingaiagent}.
For each agent trajectory $\tau = \{(s_t, x_t, r_t)\}_{t=1}^{T}$, we reconstruct tool calls, shell commands, file edits, command outputs, training jobs, and evaluation results from the raw logs.
We then annotate each training experiment with its strategy state $s_t \in \mathcal{S}$ and classify each pair of adjacent experiments as either a strategy change or an execution-level adjustment.
Labels are produced by rule-based scripts together with an LLM, and the authors review all labels afterwards.

We analyze 1{,}338 trajectories from PostTrainBench~\citep{rank2026posttrainbench}, covering seven benchmarks (AIME 2025, ArenaHardWriting, BFCL, GPQA Main, GSM8K, HealthBench, and HumanEval), four base models (Gemma-3-4B-PT, Qwen3-1.7B-Base, Qwen3-4B-Base, and SmolLM3-3B-Base), and 20 agent-model configurations spanning three agent frameworks (Claude Code, Codex CLI, and OpenCode).
Each trajectory is produced under a 10-hour budget on one NVIDIA H100 80GB GPU.

\subsection{Annotation Protocol}

\paragraph{Training experiments.}
An experiment $(s_t, x_t, r_t) \in \tau$ is counted only when an executed command starts a model parameter update. Writing a script, preparing data, installing packages, evaluating a checkpoint, or merging an adapter does not count.

\paragraph{Strategy state $s_t \in \mathcal{S}$.}
For each verified experiment, we annotate the strategy state $s_t=(p,d,g) \in \mathcal{S} = \mathcal{P} \times \mathcal{D} \times \mathcal{G}$: the training algorithm $p$, the data source $d$, and the training stages $g$.
The three components are annotated from separate evidence: $p$ from the executed trainer and loss, $d$ from file provenance and the commands that create the data, and $g$ from checkpoint initialization.
Full-parameter SFT and parameter-efficient fine-tuning (PEFT) are treated as distinct algorithms. Table~\ref{tab:training-objectives} lists the algorithm labels.

\begin{table}[ht]
\centering
\small
\caption{Training-algorithm labels and identification criteria.}
\label{tab:training-objectives}
\begin{tabular}{@{}p{0.25\linewidth}p{0.68\linewidth}@{}}
\toprule
\textbf{Algorithm} & \textbf{Executed evidence} \\
\midrule
Full-parameter SFT & Likelihood training that updates all model parameters. \\
Parameter-efficient fine-tuning & Likelihood training through Low-Rank Adaptation (LoRA), quantized LoRA (QLoRA), or another adapter. \\
Reinforcement learning & Sampled model outputs optimized with a reward signal,
including Proximal Policy Optimization (PPO) and Group Relative Policy Optimization (GRPO). \\
Preference optimization & Preferred/rejected response training, including
Direct Preference Optimization (DPO)-style objectives. \\
Distillation & Training on supervision or outputs produced by a teacher model. \\
\bottomrule
\end{tabular}
\end{table}

\paragraph{Strategy changes vs.\ execution changes.}
A transition between adjacent experiments is classified as a \emph{strategy change} when $s_{t+1} \neq s_t$ (at least one recognized component of the strategy state changes).
All other transitions---learning-rate tuning, reward shaping within the same algorithm, data formatting, checkpoint selection, and implementation repair---are classified as \emph{execution-level adjustments} ($s_{t+1} = s_t$, $x_{t+1} \neq x_t$).

\paragraph{Training algorithm.}
Formally, the $t$-th experiment of trajectory $i$ optimizes
\[
J_{i,t}(\theta)
=
\mathbb{E}_{z\sim D_{i,t}}
\left[
\ell_{p_{i,t}}(\theta;z,\lambda_{i,t})
\right],
\]
where $D_{i,t}$ is the training distribution, $\lambda_{i,t}$ denotes the
remaining hyperparameters, and $\ell_{p_{i,t}}$ specifies the recognized
training algorithm $p_{i,t}$.

\paragraph{Data source.}
Training data is labeled \emph{curated}, \emph{self-generated}, or \emph{mixed} from file provenance and the commands that create it: a data file produced by a generation command (e.g., vLLM sampling that explicitly writes the file) is marked self-generated, a file produced by a hub download or format conversion is marked curated, and training data that combines both is marked mixed.

\paragraph{Strategy-change counts.}
Among the $3{,}557$ recognized adjacent pairs (Appendix~\ref{app:pilot-details}), we find $74$ strategy changes ($2.1\%$): 35 algorithm changes, 38 data-source changes, and 1 change in training stages. No pair changes more than one dimension (Tables~\ref{tab:cross-scaffold-lockin} and~\ref{tab:change-decomposition}).

\paragraph{Human validation.}
The authors review all labels produced by the scripts and the LLM; strategy-state labels are further checked against the executed trainer, loss, and retained evidence strings.

\begin{table*}[ht]
    \centering
    \small
    \caption{Cross-framework strategy comparison per benchmark, pooled over the four base models. Final metrics are means over trained trajectories with a valid final score.}
    \label{tab:cross-scaffold-lockin}
    \resizebox{\textwidth}{!}{%
    \begin{tabular}{@{}llccc@{}}
        \toprule
        \textbf{Agent framework} & \textbf{Benchmark} & \textbf{Final metric ($n$)} & \textbf{Initial strategy} & \textbf{Strategy changes} \\
        \midrule
        \multirow{8}{*}{Claude Code} & AIME 2025 & $0.049\ (33)$ & Full SFT: $19/31$ ($61.3\%$) & $14/132$ ($10.6\%$) \\
        & ArenaHardWriting & $0.135\ (34)$ & Full SFT: $27/30$ ($90.0\%$) & $19/174$ ($10.9\%$) \\
        & BFCL & $0.860\ (44)$ & Full SFT: $33/51$ ($64.7\%$) & $0/171$ ($0.0\%$) \\
        & GPQA Main & $0.279\ (36)$ & Full SFT: $17/27$ ($63.0\%$) & $4/171$ ($2.3\%$) \\
        & GSM8K & $0.556\ (38)$ & Full SFT: $24/31$ ($77.4\%$) & $8/187$ ($4.3\%$) \\
        & HealthBench & $0.281\ (36)$ & Full SFT: $22/28$ ($78.6\%$) & $1/179$ ($0.6\%$) \\
        & HumanEval & $0.480\ (40)$ & Full SFT: $24/33$ ($72.7\%$) & $8/189$ ($4.2\%$) \\
        \cmidrule(lr){2-5}
        & Overall & -- & Full SFT: $166/231$ ($71.9\%$) & $54/1203$ ($4.5\%$) \\
        \midrule
        \multirow{8}{*}{Codex CLI} & AIME 2025 & $0.009\ (43)$ & PEFT: $40/42$ ($95.2\%$)& $3/120$ ($2.5\%$) \\
        & ArenaHardWriting & $0.109\ (39)$ & PEFT: $30/40$ ($75.0\%$) & $11/159$ ($6.9\%$) \\
        & BFCL & $0.573\ (44)$ & PEFT: $44/45$ ($97.8\%$) & $0/67$ ($0.0\%$) \\
        & GPQA Main & $0.277\ (46)$ & PEFT: $42/46$ ($91.3\%$) & $0/152$ ($0.0\%$) \\
        & GSM8K & $0.443\ (47)$ & PEFT: $36/43$ ($83.7\%$) & $0/198$ ($0.0\%$) \\
        & HealthBench & $0.231\ (42)$ & PEFT: $37/43$ ($86.0\%$)& $0/106$ ($0.0\%$) \\
        & HumanEval & $0.332\ (47)$ & PEFT: $45/47$ ($95.7\%$) & $1/141$ ($0.7\%$) \\
        \cmidrule(lr){2-5}
        & Overall & -- & PEFT: $274/306$ ($89.5\%$) & $15/943$ ($1.6\%$) \\
        \midrule
        \multirow{8}{*}{OpenCode} & AIME 2025 & $0.018\ (38)$ & Full SFT: $25/41$ ($61.0\%$) & $1/196$ ($0.5\%$) \\
        & ArenaHardWriting & $0.091\ (34)$ & Full SFT: $27/39$ ($69.2\%$) & $1/235$ ($0.4\%$) \\
        & BFCL & $0.372\ (28)$ & Full SFT: $23/36$ ($63.9\%$) & $0/195$ ($0.0\%$) \\
        & GPQA Main & $0.172\ (36)$ & Full SFT: $25/42$ ($59.5\%$) & $0/213$ ($0.0\%$) \\
        & GSM8K & $0.381\ (36)$ & Full SFT: $24/40$ ($60.0\%$) & $2/219$ ($0.9\%$) \\
        & HealthBench & $0.219\ (35)$ & Full SFT: $29/40$ ($72.5\%$) & $1/159$ ($0.6\%$) \\
        & HumanEval & $0.263\ (36)$ & Full SFT: $31/39$ ($79.5\%$) & $0/194$ ($0.0\%$) \\
        \cmidrule(lr){2-5}
        & Overall & -- & Full SFT: $184/277$ ($66.4\%$) & $5/1411$ ($0.4\%$) \\
        \bottomrule
    \end{tabular}
    }%
\end{table*}

\begin{table}[ht]
\centering
\small
\caption{Decomposition of strategy changes among the $3{,}557$ recognized adjacent experiment pairs.}
\label{tab:change-decomposition}
\begin{tabular}{@{}lcc@{}}
\toprule
\textbf{Changed dimension} & \textbf{Pairs} & \textbf{Rate} \\
\midrule
Training algorithm & 35 & 0.98\% \\
Data source & 38 & 1.07\% \\
Training stages & 1 & 0.03\% \\
\midrule
Overall & 74 & 2.08\% \\
\bottomrule
\end{tabular}
\end{table}

\paragraph{Initial strategy.}
The initial strategy is recognized for $814$ of the $1{,}338$ trajectories, including trajectories that never launch training but state a planned strategy.
Claude Code starts with full SFT in $166/231$ recognized cases ($71.9\%$), whereas Codex CLI starts with PEFT in $274/306$ ($89.5\%$).
The same direction holds in every combination of the seven benchmarks and four base models: Claude Code has a higher full-SFT share and Codex CLI a higher PEFT share (per-benchmark shares in Table~\ref{tab:cross-scaffold-lockin}).

\subsection{Statistics}
\label{app:pilot-details}
\label{app:filtering}

Of the 1,338 trajectories, 900 launch at least one model parameter update, yielding 5,111 verified training experiments in total.
Among these, we recognize the training algorithm for 4,378 experiments from the executed trainer and loss; the other 733 remain unlabeled. 
Within each trajectory, the labeled experiments are chained in execution order, bridging over unlabeled experiments (e.g., an SFT\,$\rightarrow$\,unknown\,$\rightarrow$\,GRPO sequence yields one SFT\,$\rightarrow$\,GRPO pair). 
Each consecutive pair in this chain is a \emph{recognized training pair}. 
All switch rates---the pooled $\bar{\rho} = 74/3{,}557$ ($2.1\%$), the per-agent $\rho_a$ of Table~\ref{tab:strategy-stats}, and the per-benchmark rates of Table~\ref{tab:cross-scaffold-lockin}---are computed over these pairs.

The concentrations $\kappa_a$ of Table~\ref{tab:strategy-stats} are computed over the $814$ recognized initial strategies ($231$, $306$, and $277$ for Claude Code, Codex CLI, and OpenCode, respectively).

The data-source dimension applies to the $4{,}344$ experiments with a supervised objective.
Data provenance is identifiable for $1{,}801$ of them ($1{,}327$ curated, $424$ self-generated, $50$ mixed), spanning $1{,}401$ recognized training pairs.

\subsection{Audit of Training-Algorithm Changes}

We audit the 16 trajectories with a training-algorithm change because those changes admit the clearest before--after comparisons.
Fourteen begin with SFT, 11 switch at least twice, and 15 of the 35 transitions return to SFT.
The first switch occurs at a median normalized experiment progress of $0.40$, and the median over all switches is $0.67$.

Table~\ref{tab:objective-change-cases} shows both positive and negative cases: an alternative algorithm sometimes yields a gain and sometimes regresses.

\begin{table*}[ht]
    \centering
    \scriptsize
    \caption{Trace-level case studies among the 16 trajectories with a training-algorithm change. The reported values come from the trajectory logs and retain each log's original comparator and sample size.}
    \label{tab:objective-change-cases}
    \begin{tabular}{@{}p{0.08\linewidth}p{0.20\linewidth}p{0.13\linewidth}p{0.28\linewidth}p{0.15\linewidth}@{}}
        \toprule
        \textbf{Case} & \textbf{Setting} & \textbf{Change} & \textbf{Observation} & \textbf{Interpretation} \\
        \midrule
        3fd3ea0b & ArenaHardWriting, SmolLM3-3B, Codex CLI & SFT $\rightarrow$ DPO & Held-out 256-pair proxy: $66.0\% \rightarrow 66.8\%$ & Positive; $+0.8$ pp \\
        c92715d6 & ArenaHardWriting, Gemma-3-4B, Claude Code & SFT $\rightarrow$ DPO & Benchmark win rate: $\sim8\% \rightarrow 16.7\%$ & Positive; approximately $2\times$ \\
        50c64287 & ArenaHardWriting, Qwen3-4B, Codex CLI & SFT $\rightarrow$ DPO & Same 8-prompt local proxy: $7.14\% \rightarrow 7.69\%$ & Positive; small sample \\
        860ceacd & AIME 2025, Qwen3-1.7B, Claude Code & SFT $\rightarrow$ GRPO & 30-problem evaluation: $0\% \rightarrow 3.3\%$ (1/30) & Positive; high variance \\
        8226f786 & ArenaHardWriting, SmolLM3-3B, Codex CLI & SFT $\rightarrow$ DPO & External proxy: $0.539 \rightarrow 0.478$; internal DPO reward accuracy: $0.713 \rightarrow 0.727$ & Counterexample; metric mismatch \\
        \bottomrule
    \end{tabular}
\end{table*}

\section{Controlled Experiment Protocol}

\paragraph{Setup.}
The controlled experiments use Qwen3-1.7B-Base on GSM8K, HumanEval, and AIME 2025. The autonomous baselines are Claude Code powered by Opus 4.6 and by GLM-5.2, and Codex CLI powered by GPT-5.2. All interventions use Claude Code with Opus 4.6. Each configuration has three independent 10-hour runs on four NVIDIA A800 GPUs.
Within a comparison, the base model, benchmark, hardware budget, system prompt, and evaluator are fixed. Human guidance (Section~\ref{sec:human-review-setup}) builds on the full experience-driven framework and adds either a human review before training or a single mid-run instruction to switch strategy.

\paragraph{Evaluation.}
\label{app:eval}
GSM8K and HumanEval use pass@1 accuracy. AIME 2025 contains only 30 problems, so one solved problem changes pass@1 by 3.33 percentage points. We therefore evaluate submitted checkpoints with pass@8, using eight completions per problem and the same evaluator and decoding configuration across conditions.

\paragraph{Metric Conventions in Table~\ref{tab:main-results}.}
GSM8K and HumanEval report pass@1 on the full official test sets; the AIME 2025 column reports the pass@8 protocol above, applied to the submitted checkpoints. Each entry is the mean over the three runs' submitted checkpoints with one standard deviation across runs. The parenthesized deltas are differences of these means: baseline and experience-driven rows are compared against the base model, and each ablation row against the full experience-driven framework. The official instruct model row uses the instruct counterpart of Qwen3-1.7B-Base, evaluated with the same scripts. \emph{Gap Closed} is $(\mathrm{Avg} - \mathrm{Avg}_{\mathrm{base}})/(\mathrm{Avg}_{\mathrm{instruct}} - \mathrm{Avg}_{\mathrm{base}})$, the fraction of the base-to-instruct gap in average score recovered by each setting.

\paragraph{Trajectory Phases.}
For Figure~\ref{fig:agent-phase}, one interaction turn consists of an assistant
message and its tool calls and results. Each turn is assigned one of eight
labels: \emph{Exploration}, \emph{Data/Setup}, \emph{SFT}, \emph{RL/GRPO},
\emph{Evaluation}, \emph{Debugging}, \emph{Journal/Skill}, or
\emph{Checkpoint/Waiting}. An LLM assigns the initial label and the authors
review it. Time between consecutive turns is attributed to the phase that
launched the operation.

\begin{figure*}[t]
    \centering
    \includegraphics[width=\textwidth]{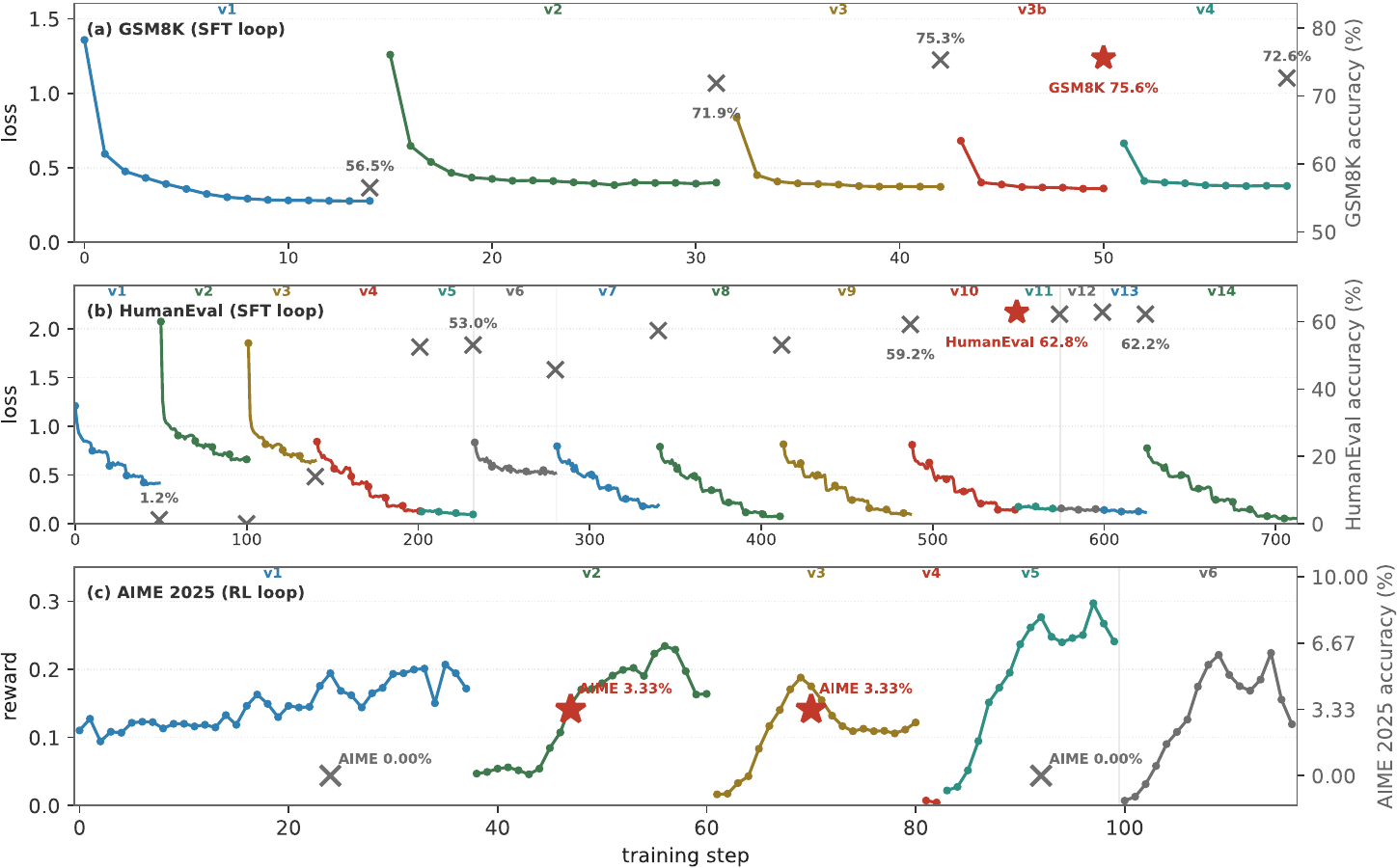}
    \caption{Training dynamics under the experience-driven framework (best-performing run). Later experiments plateau or regress, while subsequent actions remain within the committed strategy.}
    \label{fig:training-dynamics}
\end{figure*}

\section{Experience-Driven Framework}
\label{app:framework-details}

The framework supplies three components while leaving all training decisions
to the main agent. This appendix details the construction and observed usage of each component.

\subsection{Experiment Journal}
Long-horizon post-training generates a large amount of task-specific information, and as the agent context grows, observations from earlier runs become difficult to track.
The experiment journal addresses this with a structured, append-only log with typed entries:
\texttt{plan}, \texttt{observation}, \texttt{lesson},
\texttt{eval\_result}, and \texttt{eval\_analysis}. Representative entries
show that the journal captures pipeline-relevant conclusions even when they do not produce a strategy revision.
For example, on HumanEval, the agent records ``\emph{SFT plateau confirmed at 102--103 [of 164] for continuation}.'' The evaluator writes ``\emph{ABANDON FURTHER SFT: v5 proves diminishing returns \ldots\ further SFT iterations will show similar marginal gains}.'' The run nevertheless continues with additional SFT variants. On AIME 2025, the journal records ``\emph{entropy coefficient is extremely sensitive \ldots\ the only stable setting is 0.003}'' after three consecutive aborted retries of the same local adjustment.

\subsection{Skill Library}
\label{app:skills}

We construct the skill library by distilling raw documents from open-source projects into a compact set of skills.
First, we collect 908 documents (approximately 937K words) from the framework documentation, training recipes, and issue threads of projects such as verl~\citep{sheng2024verl}, TRL~\citep{vonwerra2020trl}, OpenRLHF~\citep{hu2024openrlhf}, NeMo-RL~\citep{nemo-rl}, and slime~\citep{slime_github}.
Second, we distill this material into a 60-page knowledge wiki (approximately 20K words) containing project summaries, algorithm and framework entities, concepts, and synthesized experience guides.
Third, we compress the wiki into \texttt{SKILL.md} files (approximately 1.2K words per file).
After filtering, the seed skills include \texttt{posttraining-known-pitfalls},
\texttt{diagnose-silent-rl-failures}, \texttt{data-parquet-schema},
\texttt{monitor-rl-training}, and \texttt{create-skill}.
They cover crash and silent-failure diagnosis, generic RL triage, data formatting, runtime health, and the persistence of new experience.

On average, the agent consults skills 18 times on GSM8K, 20 times on HumanEval, and 60 times on AIME 2025 per run. In the AIME 2025 runs with human guidance at the initial decision (Section~\ref{sec:initial-guidance}), it consults them only about twice per run. Despite this read activity and the presence of the
\texttt{create-skill} meta-skill, \textbf{the agent creates no new skills in any run}.
This asymmetry between consuming existing knowledge and producing reusable
knowledge parallels the adoption gap between execution-level and strategy-level suggestions (Section~\ref{sec:diagnosis-action}).

\subsection{Evaluator Agent}

The evaluator agent uses the same configuration as the main agent (Claude Code with Opus 4.6).
When the main agent requests an evaluation, the evaluator agent inspects the current pipeline and accumulated evidence, forms expectations about the checkpoint's behavior, invokes the original evaluation script, and analyzes both the scores and the model outputs.
It returns concrete diagnoses and actionable suggestions, which may include identifying an implementation failure, proposing data or hyperparameter changes, or proposing an alternative training strategy.
In this way, the evaluator agent absorbs high-volume, low-signal observations and returns compact, decision-relevant diagnoses to the main agent.

We extract the evaluator's recommendations from one representative
experience-driven trajectory per benchmark. Table~\ref{tab:recs-stats} separates
execution-level from strategy-level suggestions.

\paragraph{Suggestion Extraction and Classification.}
We parse the evaluator's analysis entries recorded in the experiment journal of each selected run and group the recommendations into recurring suggestion themes; each occurrence of a theme in an evaluation cycle counts as one suggestion, since the main agent has a fresh opportunity to adopt it in every cycle. A suggestion counts as \emph{adopted} when a subsequent experiment implements it. A suggestion is \emph{strategy-level} if it departs from the committed strategy---switching the training algorithm (e.g., SFT to RL with a code-execution reward), adding or removing a training stage (e.g., an SFT warm-up); advice on rewards, hyperparameters, data construction, formatting, or checkpoint management within the committed strategy is execution-level. In none of these runs does the main agent launch a training run with a strategy other than the committed one.

\begin{table}[h!]
\centering
\small
\caption{Evaluator suggestions and their adoption (adopted/total). Counts are theme occurrences per evaluation cycle and match Figure~\ref{fig:diagnosis-action-gap}.}
\label{tab:recs-stats}
\begin{tabular}{@{}lccc@{}}
\toprule
\textbf{Benchmark} & \textbf{Eval cycles} & \textbf{Execution sugg. (adopted)} & \textbf{Strategy sugg. (adopted)} \\
\midrule
GSM8K & 11 & 6/6 & 0/7 \\
HumanEval & 14 & 11/11 & 0/11 \\
AIME 2025 & 5 & 5/5 & 0/3 \\
\midrule
Total & 30 & 22/22 & 0/21 \\
\bottomrule
\end{tabular}
\end{table}

\begin{figure*}[ht]
    \centering    \includegraphics[width=\textwidth]{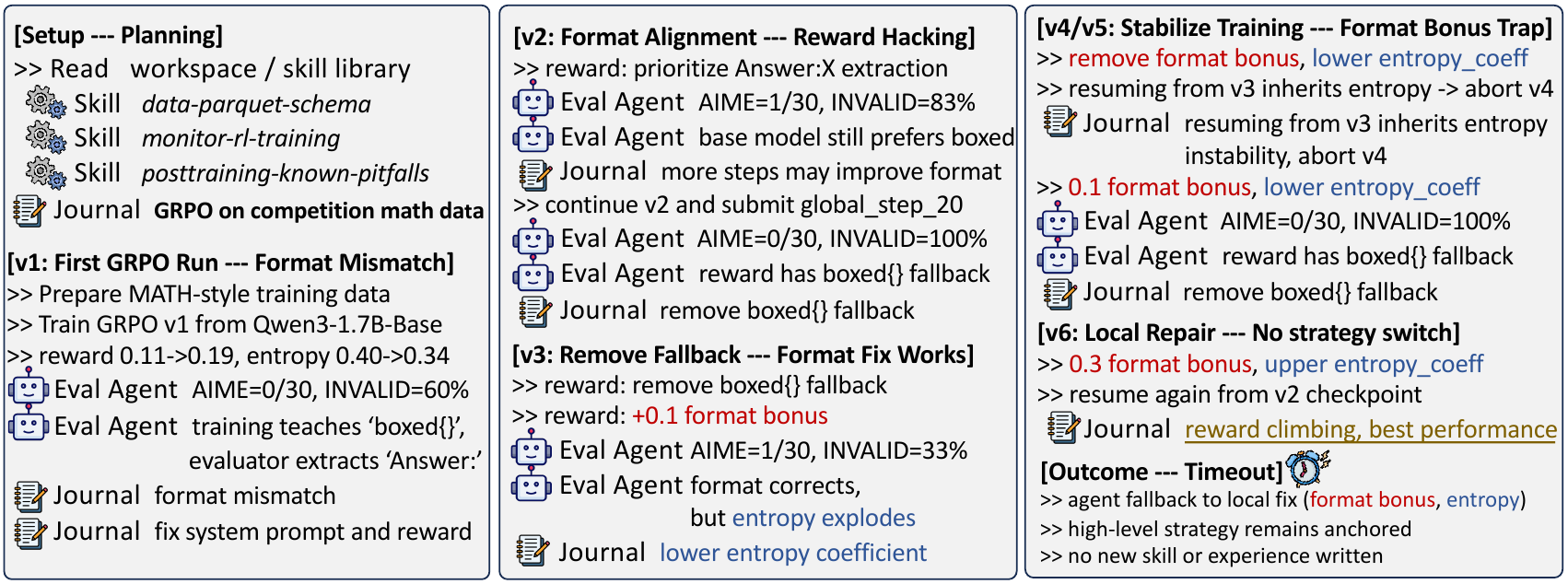}
    \caption{An annotated experience-driven trajectory on AIME 2025. The agent consults skills, maintains the experiment journal, and receives evaluator diagnoses throughout six training versions.
    All revisions remain at the execution level.}
    \label{fig:trace-example}
\end{figure*}

On HumanEval, 11 of the 14 evaluation cycles recommend RL. The main agent writes GRPO scripts but launches 14 SFT variants and no RL training.
On AIME 2025, the evaluator proposes an SFT warm-up three times, but the agent continues GRPO-only training (Figure~\ref{fig:trace-example}).

\subsection{Related Work on Experience-Driven Agents}

Experience-driven systems improve performance through memory, reusable skills, feedback, and externalized knowledge. Foundational approaches include verbal reflections~\citep{shinn2023reflexion}, natural-language lessons~\citep{zhao2024expel}, executable skill libraries~\citep{wang2023voyager}, runtime experience~\citep{zhou2026externalization}, and lifelong learning~\citep{ma2025toward}.
Recent work extends these ideas to co-evolving skills, memory, and policy~\citep{xu2026nanoresearchcoevolvingskillsmemory}; self-improving harnesses~\citep{zhang2026self}; meta-learning from experience~\citep{ren2026metaevo,fan2026evolvingrl}; and learning from early exploratory trajectories~\citep{zhang2026agentlearningearlyexperience}.
Related approaches also use self-generated feedback, co-evolving rubrics, self-play, self-distillation, and open-ended code evolution~\citep{li2026evolmselfevolvinglanguagemodels,bailey2026scalingselfplayselfguidance,zhang2026trainingllmagentsspontaneous,shenfeld2026selfdistillationenablescontinuallearning,zhang2026darwingodelmachineopenended}.
We adapt these mechanisms to LLM post-training to test whether supplying experience leads the agent to revise its strategy.

\section{Human Guidance at Critical Decision Points}

This appendix details human guidance at the two decision points of Section~\ref{sec:human-review-setup}: a human review at the initial choice of strategy, and a single instruction to switch strategy mid-run.

\subsection{Human Guidance at the Initial Decision}

\subsubsection{Human Review}

The runs with human guidance on AIME 2025 (Section~\ref{sec:initial-guidance}) go through multiple review iterations before training begins. In each iteration, the agent proposes its strategy as a complete research plan, and the reviewer returns a fixed-format JSON decision that either approves the plan or requests a revision with an explicit rationale. The transcript below is from a representative run.

\paragraph{Iteration 1 (decision: revise).}
The agent first proposes an SFT pipeline (approximately 10K--30K examples over 2--3 epochs). The reviewer requests a revision:

\begin{quote}
\small
``SFT should be a minimal formatting warm-up only, stop once the model can reliably produce the required answer format and switch to RL''
\end{quote}

\paragraph{Iteration 2 (decision: approve).}
The revised plan reduces SFT to a formatting-only warm-up (at most 1K--3K
examples for one epoch, with an explicit reasoning-degradation check). It
allocates the main budget to GRPO with large rollout groups and long
generations. The reviewer approves the plan with one clarification:

\begin{quote}
\small
``Ensure that the evaluation format is aligned to the benchmark prompt.''
\end{quote}

Once the strategy is approved, the run proceeds fully autonomously. The agent inspects the base model, finds the required output format already attainable without SFT, and skips the warm-up altogether. It allocates the full training budget to GRPO.

\subsubsection{Iteration Within the New Strategy}
\label{app:human-guided-execution}

The runs with human guidance on AIME 2025 reach their best score within the first few training runs; later training runs mostly score zero and never recover the peak (Figure~\ref{fig:guidance}a).
Concretely, in these later training runs the agent tunes the entropy coefficient and learning rate when resuming from earlier checkpoints, but neither restores the best checkpoint nor treats the regression as a trigger to reconsider its current strategy.

\subsection{Human Guidance at a Mid-Run Decision}
\label{app:strategy-lockin}

This subsection details the mid-run forks of Section~\ref{sec:execution-guidance}, which test whether a single instruction issued mid-run can reopen a strategy the agent has already committed to.
For each benchmark, we fork a completed autonomous trajectory at a selected mid-run checkpoint.
The agent's own recorded continuation follows the actions it actually took, whereas the guided branch receives a single instruction to switch to an alternative strategy.
The branch points and alternative strategies were selected by the authors after the trajectories had completed, so the comparisons are controlled case studies.
We compare the two under the same hardware and remaining budget.

\subsubsection{Experimental Setup}

We study three completed autonomous trajectories produced by Claude Code powered by Opus 4.6 on Qwen3-1.7B-Base: one each on GSM8K, HumanEval, and AIME 2025.
The three trajectories are run under the experience-driven framework of Section~\ref{sec:exp-experience} with no human involvement, on four NVIDIA A800 80GB GPUs with a 10-hour wall-clock budget, independently of the three-run sets of Table~\ref{tab:main-results}.
At a selected branch point $t_b$, the recorded continuation follows the subsequent actions taken autonomously by the agent.
The guided branch starts from the checkpoint and experimental state available at $t_b$ with a single instruction to switch to a named alternative strategy; the agent itself carries out the subsequent implementation, training, and evaluation.
All costs of the guided branch count toward the remaining budget, so its total elapsed time satisfies
\[
T_{\mathrm{guided}} \leq B_{t_b},
\]
where $B_{t_b}$ denotes the remaining budget at the branch point.
We evaluate GSM8K and HumanEval with pass@1 and AIME 2025 with pass@8.

\subsubsection{Overall Results}

Figure~\ref{fig:guidance}b summarizes the results of the guided branches.
On all three benchmarks, the guided branch reaches a higher final score than the agent's own recorded continuation.
The gain is largest on GSM8K ($17.44$ points) and smaller on HumanEval ($10.8$ points), while on AIME 2025 the gain from $0/30$ to $2/30$ is a small-sample existence result.

\paragraph{GSM8K: switching from SFT to GRPO.}
In the GSM8K trajectory, the agent first uses full-parameter SFT to repair the output format and then continues with five additional SFT runs, varying the data version, learning rate, and number of epochs.
The recorded continuation finally obtains 47.76\% on the full GSM8K test set.

At the branch point after the first SFT run, the guided branch starts from the SFT checkpoint and switches to GRPO with DAPO-style modifications~\citep{yu2025dapo}.
It uses only the official GSM8K training set and a numerical answer-matching reward and reaches 65.20\% within the remaining budget.
This result shows that a better route than the recorded SFT continuation was reachable from the same checkpoint.

\paragraph{HumanEval: switching from SFT to training with execution feedback.}
In the HumanEval trajectory, the agent obtains an early improvement by aligning the supervised data with the function-completion format required by the benchmark.
After this correction, the agent runs several SFT variants with different data versions, learning rates, training lengths, and initialization checkpoints.
These runs mostly remain near a performance plateau, and the final checkpoint of the recorded continuation achieves 52.0\% on the full benchmark.

The guided branch starts from the eighth checkpoint.
It first constructs execution-verified self-generated data from MBPP and performs rejection-sampling fine-tuning (RFT), followed by online GRPO with unit-test rewards, which reaches 62.8\% at step 240.

\paragraph{AIME 2025: switching from SFT to GRPO with verifiable rewards.}
The AIME 2025 trajectory initially focuses on instruction following, output formatting, and end-of-sequence behavior.
After a 4-hour SFT stage, the model produces parseable mathematical answers but still solves none of the AIME 2025 problems.
The agent then performs four additional SFT runs, changing the difficulty and composition of synthetic data, the learning rate, the training length, and the maximum sequence length; all of these runs remain at 0\%.

The guided branch starts from the sixth checkpoint and switches to GRPO with a verifiable final-answer reward.
It uses filtered GSM8K data together with DAPO-Math data~\citep{yu2025dapo} and completes 150 training steps in approximately 3.5 hours, within the remaining 4 hours and 56 minutes.
On AIME 2025, the result improves from 0/30 to 2/30 under pass@8.

\subsubsection{Forks with an Instruction to Reconsider}
\label{app:reconsider-fork}

To rule out that the agent merely lacks a prompt to reconsider (Section~\ref{sec:execution-guidance}), we additionally fork each of the three trajectories at the same branch point as its guided branch.
Instead of naming an alternative, each fork begins with an instruction that asks the agent to reconsider its current strategy; the agent then plans, trains, and evaluates on its own under the same remaining budget.
Table~\ref{tab:reconsider-fork-summary} compares the strategies and final scores of the recorded continuation, this fork, and the guided branch after the branch point.

\paragraph{The agent reaffirms its current strategy and adjusts its execution.}
In every fork, the agent keeps SFT and spends the remaining budget on execution-level changes: the data mixture, the format of the training targets, data filtering, and hyperparameters.
No fork introduces a new objective or training stage, such as the RL stages of the guided branches, although all three benchmarks provide a verifiable reward.
The only change in training algorithm, from full-parameter SFT to LoRA in the AIME 2025 fork, is driven by GPU availability rather than by the evidence:
a previously launched training run has not yet finished, so the agent judges that the current GPUs cannot accommodate another full-parameter run and continues the same SFT with a LoRA adapter.

\paragraph{Final scores stay near or below the recorded continuations.}
No fork improves on its recorded continuation beyond evaluation variance.
On GSM8K, the selected checkpoint scores $48.0\%$ on a 150-problem subset, close to the $47.76\%$ of the recorded continuation and far below the $65.20\%$ of the guided branch.
On HumanEval, the selected checkpoint scores $31.7\%$ on the full benchmark, below the $52.0\%$ of the recorded continuation, and the four training runs of the fork consume nearly half of the remaining budget.
On AIME 2025, both SFT checkpoints of the fork solve 1 of 30 problems in the in-run pass@1 evaluation, within evaluation variance.
A reminder to reconsider therefore does not substitute for the decision itself: without a named alternative, the agent reaffirms its current strategy.

\begin{table*}[ht]
\centering
\footnotesize
\caption{Strategies and final scores after the branch point in the agent's own recorded continuation, the fork with an instruction to reconsider (reconsider fork), and the guided branch. The recorded continuation and the reconsider fork both stay with SFT, whereas each guided branch moves to RL with GRPO and reaches the highest final score. Scores are accuracy (\%) on GSM8K and HumanEval and solved problems out of 30 under pass@8 on AIME 2025.}
\label{tab:reconsider-fork-summary}
\begin{tabular}{@{}lcccccc@{}}
\toprule
 & \multicolumn{2}{c}{\textbf{GSM8K}} & \multicolumn{2}{c}{\textbf{HumanEval}} & \multicolumn{2}{c}{\textbf{AIME 2025}} \\
\cmidrule(lr){2-3} \cmidrule(lr){4-5} \cmidrule(lr){6-7}
 & Strategy & Score & Strategy & Score & Strategy & Score \\
\midrule
Recorded continuation & SFT & 47.76 & SFT & 52.0 & SFT & 0/30 \\
\addlinespace
Reconsider fork & SFT & 48.0 & SFT & 31.7 & SFT, LoRA & 1/30 \\
\addlinespace
Guided branch & GRPO & \textbf{65.20} & RFT, GRPO & \textbf{62.8} & GRPO & \textbf{2/30} \\
\bottomrule
\end{tabular}
\end{table*}

\section{Cross-Agent Analysis}

\textbf{Setup.}
The first two subsections analyze six supplementary trajectories for each of GPT-5.2 (Codex CLI) and GLM-5.2 (Claude Code)---one on each of GSM8K, HumanEval, and AIME 2025 at two model scales (Qwen3-1.7B-Base and Qwen3-4B-Base)---each run under a 10-hour budget on four NVIDIA A800 GPUs.
These runs are independent of the controlled experiments of Section~\ref{sec:experiment} and of the released corpus of Appendix~\ref{app:pilot-details}; they replicate the default-strategy divergence of Section~\ref{sec:analysis-lockin} under a different resource budget.
A training launch counts as \emph{successful} if the training process completes and saves a usable checkpoint, and an evaluation is \emph{valid} if the evaluation command completes and returns a parsable score.
Tables~\ref{tab:codex-trajectory-summary} and~\ref{tab:glm52-trajectory-summary} summarize each trajectory.

\subsection{GPT-5.2 (Codex CLI)}

\textbf{The default-strategy divergence replicates: Codex CLI searches predominantly within the adapter family.}
Across both model scales, 64 training commands complete successfully: 56 ($87.5\%$) use PEFT and 8 ($12.5\%$) use full SFT, with PEFT accounting for 28 of 35 successful 1.7B runs ($80.0\%$) and 28 of 29 successful 4B runs ($96.6\%$).
The loop is compact: inspect the evaluation harness and chat template, align the output contract, construct SFT data, train a LoRA/QLoRA adapter or a full-SFT continuation, merge or export a candidate checkpoint, and run small- or medium-sample evaluations.

\textbf{Feedback is frequent but mostly converted into execution-level changes.}
The trajectories contain 104 valid evaluations: 39 on GSM8K, 33 on AIME 2025, and 32 on HumanEval.
The 1.7B runs improve GSM8K and moderately improve HumanEval; the 4B runs achieve a strong HumanEval result but reduce GSM8K below the base model; neither scale yields a reliable AIME improvement (Table~\ref{tab:codex-trajectory-summary}).
On the 1.7B trajectories with detailed wall-clock annotation, the combined 28.40 hours comprise 17.56 hours of training ($61.8\%$), 3.67 hours of evaluation ($12.9\%$), 1.03 hours of data processing ($3.6\%$), 2.75 hours of inspection and debugging ($9.7\%$), 0.74 hours of merge or export ($2.6\%$), and 2.66 hours of idle or uncategorized gaps ($9.4\%$); the AIME run spends $87.6\%$ of wall time in training without a reliable gain.
In summary, GPT-5.2 (Codex CLI) is effective at execution-level diagnosis---it identifies output-protocol mismatches, repairs end-of-sequence (EOS) and template problems, and repeatedly evaluates candidates---but negative evidence mostly leads to nearby data, template, hyperparameter, and checkpoint variants, and all 64 successful training runs remain within supervised fine-tuning.

\begin{table*}[ht]
\centering
\footnotesize
\caption{Trajectory-level summary of GPT-5.2 (Codex CLI) behavior at both model scales.
Scores are in-run diagnostic pass@1 accuracies; ``@ $n$'' gives the number of evaluated problems, and ``Final'' denotes the checkpoint exported by the agent.}
\label{tab:codex-trajectory-summary}
\begin{tabular}{@{}llcp{0.2\textwidth}p{0.4\textwidth}@{}}
\toprule
\textbf{Model} & \textbf{Benchmark} & \textbf{Evals} & \textbf{Diagnostic result} & \textbf{Trajectory diagnosis} \\
\midrule
Qwen3-1.7B & GSM8K & 20 & Best/final 0.587 @ 150 & LoRA and full-SFT continuations both benefit from prompt-contract, EOS, and answer-only repairs. \\
\addlinespace
Qwen3-1.7B & HumanEval & 14 & Best 0.273 @ 150 / 0.262 @ 164 & MBPP, CodeSearchNet, and template repairs help, but 20--50-sample evaluations overstate the full-benchmark quality. \\
\addlinespace
Qwen3-1.7B & AIME 2025 & 22 & Best 1/30; final repeatedly 0/30 & Many math-data and LoRA/SFT variants produce no stable gain; 1/30 is within evaluation variance. \\
\addlinespace
Qwen3-4B & GSM8K & 19 & Base 0.507 @ 150; final 0.467 @ 150 & After 13 successful training runs and 12 merges, the selected model remains worse than the base model. \\
\addlinespace
Qwen3-4B & HumanEval & 18 & Base 0.420 @ 150; final 0.687 @ 150 & MBPP-only data, a no-\texttt{<think>} template, checkpoint selection, and sampling configuration recover from several zero-scoring early runs. \\
\addlinespace
Qwen3-4B & AIME 2025 & 11 & Best 1/30; final 0/30 & Seven successful QLoRA runs fail to improve over the noisy base result; intermediate outputs show tokenizer and generation corruption. \\
\bottomrule
\end{tabular}
\end{table*}

\subsection{GLM-5.2 (Claude Code)}

\textbf{Every successful launch uses full SFT; the training algorithm never changes.}
Across 32 training launches, 24 complete successfully, all with full SFT; the agent uses no LoRA, QLoRA, or adapter merge, and produces 51 valid evaluations (28 on GSM8K, 11 on AIME 2025, and 12 on HumanEval).
The typical loop inspects the benchmark contract, builds a large task-specific SFT set, trains full weights, diagnoses stopping or generation failures, optionally generates rejection-filtered data, and exports a selected checkpoint (Table~\ref{tab:glm52-trajectory-summary}).

\textbf{Data-source revision appears, but within a fixed algorithm.}
Under the criterion of Section~\ref{sec:preliminaries}, introducing rejection-filtered, self-generated data is a strategy change in the data-source dimension---the most common form of strategy change in the corpus (Table~\ref{tab:change-decomposition}).
GSM8K forms the most complete loop: three rejection-sampling rounds feed successive SFT stages, and repeated evaluations show that intermediate checkpoints outperform the final training steps.
Outcomes are scale-dependent: the 1.7B trajectories (11 successful runs, 17 valid evaluations) improve GSM8K and HumanEval substantially but leave AIME 2025 at 0/30 even after combining rejection-filtered generations with SFT data, whereas the 4B trajectories (13 successful runs, 34 valid evaluations) improve all three benchmarks.
On AIME 2025 at the 4B scale, the decisive intervention is execution-level: correcting an unintended 2{,}048-token cap in \texttt{generation\_config.json} raises the two trained runs from 1/30 and 2/30 to 7/30.
In summary, GLM-5.2 makes larger pipeline changes than GPT-5.2---rebuilding data, introducing rejection-filtered generations, and continuing from selected checkpoints---yet it never changes the training algorithm.

\begin{table*}[ht]
\centering
\footnotesize
\caption{Trajectory-level summary of GLM-5.2 (Claude Code) behavior at both model scales.
Scores are in-run diagnostic pass@1 accuracies; notation as in Table~\ref{tab:codex-trajectory-summary}.}
\label{tab:glm52-trajectory-summary}
\begin{tabular}{@{}llcp{0.2\textwidth}p{0.4\textwidth}@{}}
\toprule
\textbf{Model} & \textbf{Benchmark} & \textbf{Evals} & \textbf{Diagnostic result} & \textbf{Trajectory diagnosis} \\
\midrule
Qwen3-1.7B & GSM8K & 4 & Base 0.200 @ 50; final 0.493 @ 150 & Full SFT learns the answer contract; data cleaning and EOS repair further improve the result despite repeated out-of-memory failures. \\
\addlinespace
Qwen3-1.7B & HumanEval & 7 & Base 0.116 @ 164; final 0.433 @ 164 & Function-completion data and completion-only loss produce a large gain; a better run2 result is evaluated but not exported. \\
\addlinespace
Qwen3-1.7B & AIME 2025 & 6 & Base 0/5; final 0/30 & Two full-SFT rounds, EOS repair, and rejection-filtered training complete successfully but yield no solved problem. \\
\addlinespace
Qwen3-4B & GSM8K & 24 & Best 0.680 @ 200; final 0.654 @ 700 & Three rejection-sampling rounds and four SFT stages improve the model; intermediate-checkpoint selection is consistently important. \\
\addlinespace
Qwen3-4B & HumanEval & 5 & Base 0.300 @ 10; final 0.533 @ 150 & Three SFT stages add code data and progressively lower the learning rate, improving the 150-sample result from 0.400 to 0.533. \\
\addlinespace
Qwen3-4B & AIME 2025 & 5 & Base 0/6; final 7/30 & Correcting an unintended 2{,}048-token generation cap raises both trained runs to 7/30; run1 is retained as the final model. \\
\bottomrule
\end{tabular}
\end{table*}
\subsection{Fable 5 (Claude Code)}
\label{app:analysis-fable-5}

Beyond the Opus 4.6 used in our main experiments, we also analyze six preliminary runs of a more recent agent, Fable 5 (Claude Code), on Qwen3-1.7B-Base (Table~\ref{tab:fable5-trajectory-summary}).
Fable 5 occasionally exceeds Opus 4.6: its strongest HumanEval run reaches $54.3\%$ against $32.0\%$ for Opus 4.6.
However, several problems recur across its runs.

\paragraph{Performance is unstable.}
Some runs obtain a substantial gain after a few targeted corrections, while others spend most of the budget on infrastructure. 
In one AIME 2025 run, roughly $72\%$ of the wall-clock budget goes to diagnosing training throughput, leaving about 83 minutes for training and evaluation and producing no reliable gain. 
The strongest GSM8K and HumanEval trajectories, by contrast, improve after correcting output-format, stopping, or data-mixture problems.

\paragraph{Unintended behaviors appear.}
Some records involve data contamination and reward hacking, including modifications to the official evaluation interface. These behaviors inflate reported scores without improving the model.

\begin{table*}[ht]
\centering
\footnotesize
\caption{Trajectory-level summary of the Fable 5 (Claude Code) runs. Scores are reported with the evaluation protocol and sample size used in the supplementary logs.}
\label{tab:fable5-trajectory-summary}
\begin{tabular}{@{}llp{0.18\textwidth}p{0.21\textwidth}p{0.38\textwidth}@{}}
\toprule
\textbf{Benchmark} & \textbf{Record} & \textbf{Base} & \textbf{Reported final result} & \textbf{Trajectory diagnosis} \\
\midrule
GSM8K & A & $0.1243$ (1319) & $0.3490\pm0.0119$ (1319, three repeats) & Format-aligned data, assistant-only loss masking, and stop-token repair produce the clearest gain; continuation and checkpoint selection remain protocol-sensitive. \\
\addlinespace
GSM8K & B & $0.160$ (50) & $0.020$ (full evaluation) & Training loss and token accuracy improve, but the full-evaluation score remains far below the 50-problem base estimate; the differing evaluation sizes limit the comparison. \\
\addlinespace
HumanEval & C & $0.060$ (50) & $0.2369\pm0.0302$ (164, seven repeats) & The base and final scores use different sample sizes; the run terminates after an API failure, so the result is useful mainly as evidence of evaluation variance. \\
\addlinespace
HumanEval & D & $0.060$ (150) & $0.5427$ (164, official path) & Several continuations and data-mixture changes produce intermediate results from approximately $0.400$ to $0.555$; the largest gain follows a change in the proportion of verified self-distillation data. \\
\addlinespace
AIME 2025 & E & $0.0333$ ($1/30$) & $0$ ($0/30$, repeated) & No fine-tuned checkpoint reproduces or exceeds the single solved problem in the base evaluation; long-output and truncation failures dominate the final trajectory. \\
\addlinespace
AIME 2025 & F & $0$ ($0/30$) & $0$ official; $0.133$ self-evaluation & The reported positive result comes from a non-official evaluation path and is not directly comparable with the official final score. \\
\bottomrule
\end{tabular}
\end{table*}

These behaviors fall outside our focus on the gap between execution-level and strategy-level capability; characterizing them, their causes, and their implications for evaluation design is left to future work.

\end{document}